\documentclass[journal]{IEEEtran}
\usepackage{cite}

\ifCLASSINFOpdf
   \usepackage[pdftex]{graphicx}
   \usepackage{subfigure}
   \graphicspath{{../pdf/}{../jpeg/}}
   \DeclareGraphicsExtensions{.pdf,.jpeg,.png}
\else
   \usepackage[dvips]{graphicx}
   \usepackage{subfigure}
   \graphicspath{{../eps/}}
   \DeclareGraphicsExtensions{.eps}
\fi

\usepackage[ruled,linesnumbered]{algorithm2e}
\usepackage{color}
\makeatletter

\newcommand{\Rmnum}[1]{\expandafter\@slowromancap\romannumeral #1@}
\makeatother

\usepackage{amsmath}
\usepackage{amsfonts}       
\usepackage{bm}
 \usepackage{algorithmic}
\usepackage{array}

\ifCLASSOPTIONcompsoc
  \usepackage[caption=false,font=normalsize,labelfont=sf,textfont=sf]{subfig}
\else
  \usepackage[caption=false,font=footnotesize]{subfig}
\fi
\usepackage{fixltx2e}

\usepackage{stfloats}

\usepackage{multirow}
\begin{document}
%
\title{A Bioinspired Retinal Neural Network for Accurately Extracting Small-Target Motion Information in Cluttered Backgrounds}
%
%
%

\author{Xiao~Huang, Hong~Qiao,~\IEEEmembership{Fellow,~IEEE}, Hui~Li, and~Zhihong~Jiang
\thanks{X. Huang, H. Li and Z. H. Jiang are with the School of Mechatronical Engineering, Advanced Innovation Center for Intelligent Robots and Systems, Key Laboratory of Biomimetic Robots and Systems of Chinese Ministry of Education, Beijing Institute of Technology, Beijing, 100081, China (e-mail:7520200120@bit.edu.cn; lihui2011@bit.edu.cn; jiangzhihong@bit.edu.cn).}
\thanks{H. Qiao is with the State Key Laboratory of Management and Control for Complex Systems, Institute of Automation, Chinese Academy of Sciences, Beijing, 100190, China, is also with Beijing Key Laboratory of Research and Application for Robotic Intelligence of 'Hand-Eye-Brain' Interaction, Beijing, 100190 and the University of Chinese Academy of Sciences, Beijing, 100049, e-mail: (hong.qiao@ia.ac.cn).}
\thanks{This work is supported in part by the National Key Research and Development Program of China under 2018YFB1305300, 2017YFB1300200 and 2017YFB1300203, the China Postdoctoral Science Foundation under Grant 2020TQ0039, the National Natural Science Foundation of China under Grant 61733001, U2013602, 61873039, U1913211 and U1713215. (Corresponding author: Hui Li, Zhihong Jiang)}
}

\maketitle


\begin{abstract}
Robust and accurate detection of small moving targets in cluttered moving backgrounds is a significant and challenging problem for robotic visual systems to perform search and tracking tasks. Inspired by the neural circuitry of elementary motion vision in the mammalian retina, this paper proposes a bioinspired retinal neural network based on a new neurodynamics-based temporal filtering and multiform 2-D spatial Gabor filtering. This model can estimate motion direction accurately via only two perpendicular spatiotemporal filtering signals, and respond to small targets of different sizes and velocities through adjusting the dendrite field size of spatial filter. Meanwhile, an algorithm of directionally selective inhibition is proposed to suppress the target-like features in the moving background, which can reduce the influence of background motion effectively. Extensive synthetic and real-data experiments show that the proposed model works stably for small targets of a wider size and velocity range, and has better detection performance than other bioinspired models. Additionally, it can also extract the information of motion direction and motion energy accurately and rapidly.

\end{abstract}


%
\IEEEpeerreviewmaketitle

\section{INTRODUCTION}
%
%
%
%

\IEEEPARstart{R}{apidly} detecting small moving targets in the cluttered background and judging their motion directions, is very important for most animals to track prey and escape predators. It is also a significant and challenging problem for robotic visual systems to perform search and rescue, traffic monitoring, and any task that requires detection of small moving targets. However, it is difficult to detect these kind of targets since most shape, color, texture and structure information is lost in few pixels. Meanwhile, these targets are easily buried in the background and are indistinguishable from noise such that it becomes difficult to detect small targets under such low signal-to-clutter ratio (SCR) backgrounds. Thus, the aim of this paper is to detect small moving targets quickly and accurately in the natural cluttered environment, and extract more motion information like motion direction and energy at the same time.

For motion detection, most of conventional computer-vision solutions are for normal-sized objects like pedestrians and vehicles, including background subtraction \cite{huang2013radial,yong2018robust}, temporal differencing\cite{li2016rotation,shuigen2009motion}, statistical model\cite{tsai2009independent,woo2010environmentally} and optical flow \cite{fortun2015optical,wei2011motion}. Recent years have also seen many deep learning-based target detection methods, such as R-CNN framework \cite{girshick2016region, girshick2015fast, ren2015faster}, SSD \cite{liu2016ssd} and YOLO framework \cite{redmon2018yolov3,redmon2017yolo9000,redmon2016you}. While conventional solutions to small target detection mainly focus on infrared images\cite{gao2013infrared,bai2018derivative,dong2014a}. For instance, Gao et al.\cite{gao2013infrared} built an infrared patch-image model and transformed the small target motion detection into an optimization problem, which obtained better detection performance for different target sizes and SCR values. Bai et al.\cite{bai2018derivative} proposed the derivative entropy-based contrast measure to enhance the infrared small target and suppress background clutter. Lin et al. \cite{lin2018using} used a deep convolutional neural network (CNN) to extract small target features and suppress clutters in an end-to-end training manner, which can improve the SCR value and detection performance effectively. However, These infrared-based approaches rely on the strong temperature difference between the background and the targets, and the environment is usually required to be clean. For natural images, Sun et al. \cite{sun2018sg} developed a video eye fixation detection model using a CNN to improve the saliency detection performance. Zhu et al. \cite{zhu2020moving} recently proposed a CNN-based method called YOLOv3-SOD for detecting small targets in natural images. However, most of deep learning-based methods usually need a lot of training data and manual annotation for good performance.

Recently, many bioinspired models have emerged to perform this kind of motion detection. In nature, insects have excellent ability to search for moving features in the distance, which derives from a special class of small target motion-detecting neurons (STMDs) \cite{nordstrom2006small,barnett2007retinotopic}. Inspired by visual system in insects, some studies recently have developed several efficient models for elementary visual perception in cluttered environments. For instance, the study in \cite{wiederman2008model} firstly proposed a computational model for target discrimination, which was well matched to the properties of STMDs from a physiological perspective and detected most targets from the background successfully. After that, \cite{bagheri2017autonomous,bagheri2017performance,bagheri2015properties} improved this model further by using a facilitation mechanism and motion direction detection system, but motion direction is only divided into right/up and left/down. Similarly, \cite{wang2020a} built a computational model of directionally selective small target motion detection (DSTMD) to not only perform target discrimination, but also perceive the motion direction. However, these models generally suffer from some performance loss of detection since they were just sensitive to a small range of target sizes and moving velocities and could not suppress the movement of background effectively. The study \cite{wang2020arobust} improved the performance of DSTMD by using directional contrast information to discriminate small targets from fake features of background. But, this approach actually requires a prior determination of the motion direction of background, according to its open source code. Additionally, the study in \cite{colonnier2019bio} implemented a similar bioinpired visual processing system on a mobile robot with a biomimetic compound eye, and enabled the robot to locate and track a target like a hoverfly.


The mammalian vision system consits of low-level, intermediate-level and high-level visual processing for performing primary visual perception and complex visual cognition.  Compared with insect, the visual acuity of mammalian vision is generally higher \cite{caves2018visual}, but the temporal resolution is relatively low \cite{bostrom2016ultra}, which may be caused by a longer visual processing pathway and more complicated top-down interaction. Despite the differences, for low-level visual processing, both retinas actually have {common} neural circuit design to solve the problem of primary motion perception \cite{borst2015common,clark2016parallel}. Thus, it is attractive and significant to model mammalian low-level vision like insect-modeling mannar, and further develop higher-level visual cognition system for general machine intellegence. Currently, bioinspired studies for modelling mammalian motion vision are relatively few. A feasible solution is to use spatiotemporal energy models. For instance, Adelson and Bergen \cite{adelson1985spatiotemporal} used Gabor filters and a temporal band-pass structure to achieve elementary perception of motion, where the spatial characteristics of Gabor filter are similar to mammalian visual system. Some studies used this model to calculate the optical flow \cite{heeger1988optical}. However, such approaches generally use infinite impulse response (IIR) filters to computing the output signals, in which a phase delay is usually introduced. Another bioinspired model is the ViSTARS based on neural dynamics \cite{browning2009a,browning2009cortical}, which demonstrates how primates use motion information to segment objects and compute heading from optic flow and obstacle avoidance in response to visual inputs from realistic environments. This model is biologically plausible but with large number of dynamic parameters, which is difficult to find suitable parameters for adapting to the cluttered and dynamic environments. Meanwhile, all aforementioned approaches are mainly used for wide-field motion perception rather than small-field motion detection. 

Inspired by the neural circuitry of elementary motion vision in the mammalian retina, this paper proposes a bioinspired retinal neural network (BRNN) that can not only detect small moving targets in cluttered backgrounds, but also estimate motion direction and motion energy of targets accurately. More specifically, this model is based on the spatiotemporal energy model comprising of four neural modules corresponding to photoreceptor cells, bipolar cells, amacrine cells, and ganglion cells in the retina respectively. Compared with previous methods, the BRNN works stably for different target sizes and velocities, and has better detection performance. These advantages mainly stem from the following contributions.
\begin{enumerate}
    \item a novel spatiotemporal energy model is proposed based on a neurodynamics-based temporal filtering and multiform 2-D spatial Gabor filtering. The temporal filtering is based on cascades of leaky integrator, which can improve the neural response speed due to smaller phase delay. The motion direction can be estimated accurately by only two perpendicular direction-selective spatiotemporal filtering channels. The optimal sensing size and velocity can be mediated by adjusting the dendrite field size of antagonistic center-surround spatial filtering.
    \item An algorithm of directionally selective inhibition is proposed to suppress the target-like features in the background. If a large number of targets appear suddenly in similar directions, these target-like features will be regarded as interference targets, then the spatial filtering in these directions will be inhibited. The method can reduce the influence of background motion effectively.
\end{enumerate}

The rest of this paper is organized as follows. Section \Rmnum{2} introduces the neural mechanisms of visual motion processing in the mammalian retina. Section \Rmnum{3} provides a detailed description of BRNN for motion detection and motion information extraction. Section \Rmnum{4} describes the experimental setup and discusses the results of motion perception in synthetic and real scenarios respectively. Section \Rmnum{5} concludes this paper.

\section{VISUAL MOTION PROCESSING IN THE MAMMALIAN RETINA}
\begin{figure*}[!htbp]
    \centering
    \subfigure[]{\label{fig:s02f01a}\includegraphics[width=0.4\textwidth]{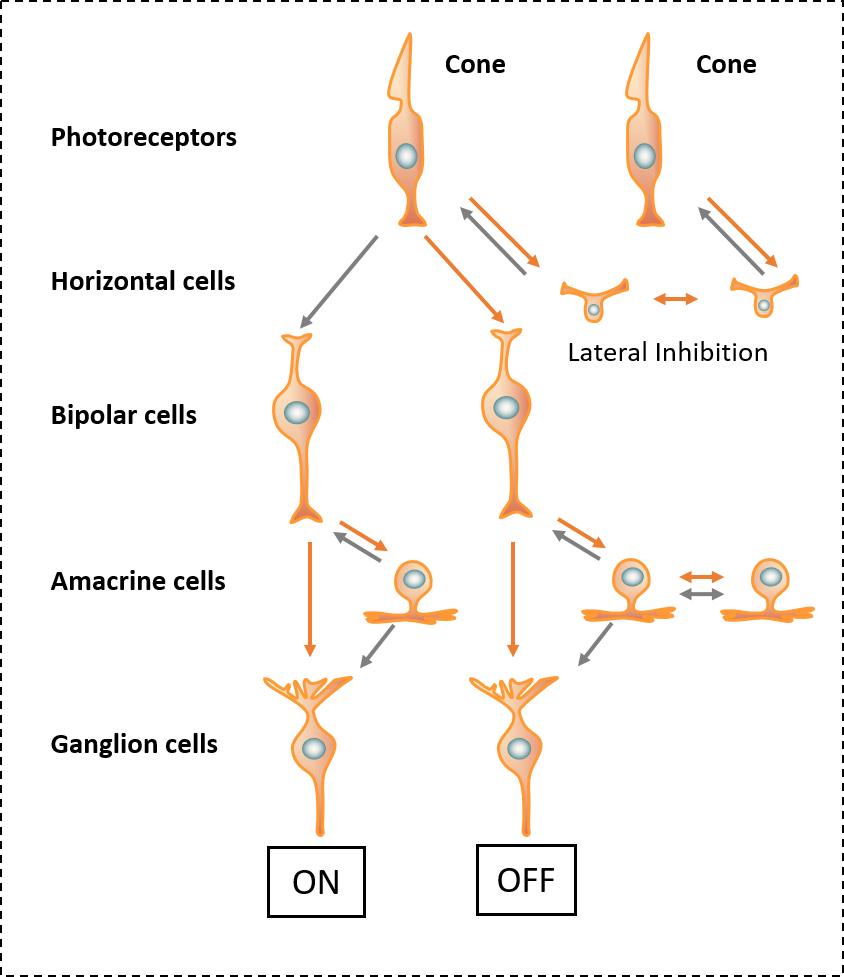}}%
    ~\hspace{15pt}
    \subfigure[]{\label{fig:s02f01b}\includegraphics[width=0.575\textwidth]{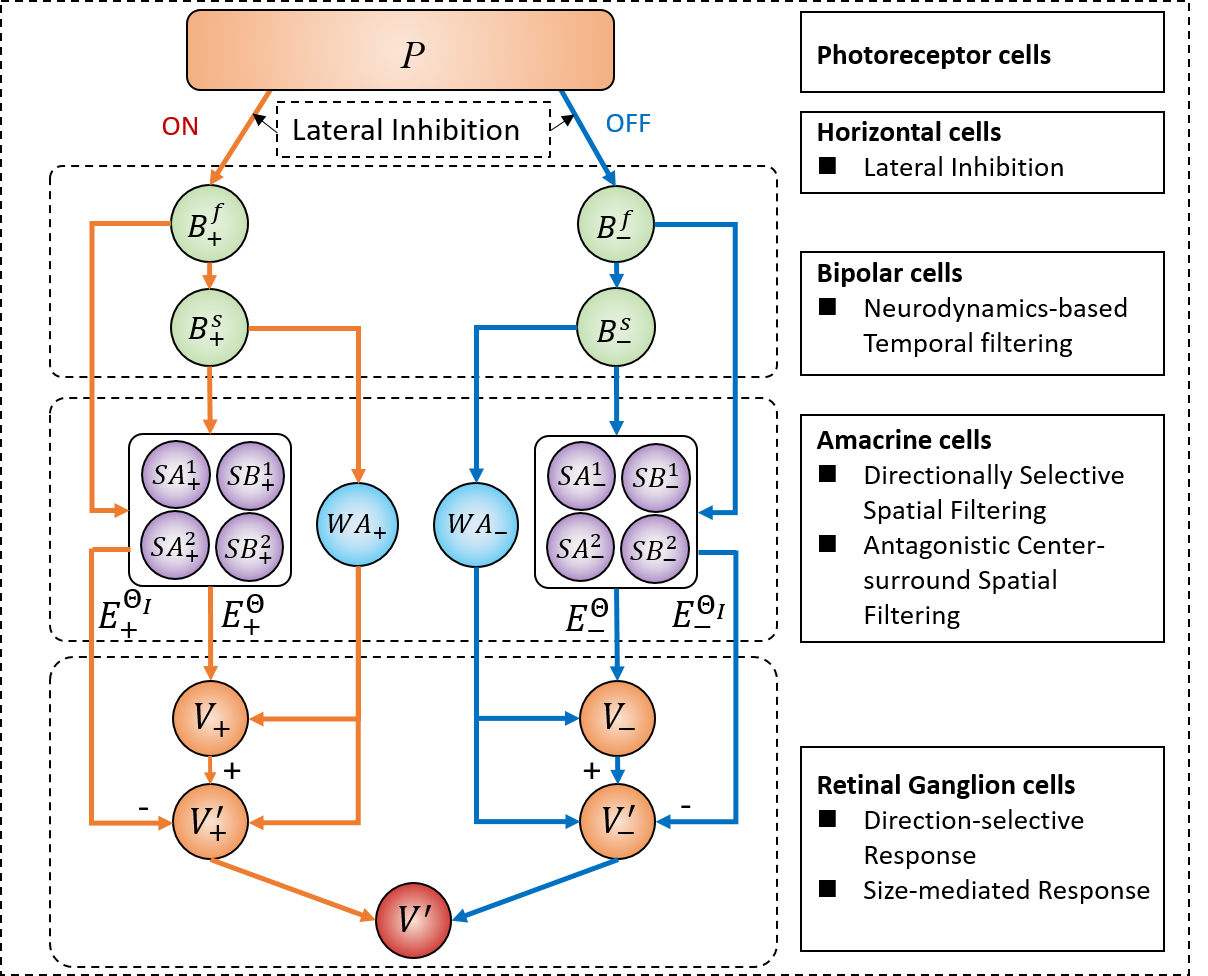}}
    \caption{The neural mechanism and computational model of visual motion processing in the mammalian retina. (a) The retinal circuitry for cone signals \cite{kendel2000principles}. Yellow arrows indicate sign-preserving connections, and gray arrows represent sign-inverting connections. (b) The computational model of primary visual motion perception.}
    \label{fig:s02f01}
\end{figure*}

The mammalian retinas generally have a layered structure that contains five principle cell types: photoreceptors, horizontal cells, bipolar cells, amacrine cells and ganglion cells \cite{masland2001fundamental}, as shown in Fig. \ref{fig:s02f01a}. First of all, there are two main types of photoreceptor, rods and cones. Rod cells, sensitive to dark light and with a poor spatial resolution, are mainly responsible for seeing at night. In daylight, cones are considered as the main source of visual processing and the density of them determines the spatial resolution of the mammalian retina. 

Next, the photoreceptor signal divides into two parallel channels, an ON and an OFF channel. The ON channel responds to contrast increments and the OFF channel responds to contrast decrements. These two pathways report complementary information and transmit more information than an ON-ON or OFF-OFF channel in motion computation \cite{clark2016parallel}. Then he the ON signals depolarize ON bipolar cells and hyperpolarize OFF bipolar cells (and vice versa) \cite{euler2014retinal}. The bipolar cells are mainly responsible for the temporal filtering of the visual information from the photoreceptors \cite{bialek1990temporal,burkhardt2007retinal}. Fast and slow responses are created based on different temporal dynamics in the bipolar layer, which is crucially important for direction selectivity in motion detection. Meanwhile the bipolar cells are also affected by the inhibitory connections from the horizontal cells.

ON and OFF bipolar cells are further connected to the amacrine cells, one type of which, called starburst amacrine cells (SACs), has been discovered to be able to produce direction-selective responses. Each of the radial SACs dendrites is sensitive to its preferred direction of motion, and converges to the SAC soma to form a specific preferred direction \cite{borst2015common}. Other directions are suppressed by the inhibitory effect of SACs dendrites. As a result, the SAC are actually responsible for the spatial filtering of the visual information. Besides direction selectivity, recent studies have found a class of wide-field amacrine cells (WACs) could endow the direction-selective circuit with an additional feature: size selectivity \cite{Hoggarth2015}. WACs contact bipolar cells and drive direct excitation and feedforward inhibition (through SACs) to ganglion cells, which allows the size selectivity to be adjusted.

Finally, retinal ganglion cells (RGCs) process the visual information from the amacrine cells and implement elementary motion perception further. Ganglion cells, as the sole output neuron of the retina, communicate visual signals to visual processing centers in the brain. There are estimated 20-30 subtypes of ganglion cells in mouse or primates \cite{dhande2015contributions}, including motion-sensitive and non-motion-sensitive ganglion cells. Therein, three groups of ganglion cells are found to be sensitive to the motion direction, ON-OFF ganglion cells, ON ganglion cells and OFF JAM-B cells \cite{borst2015common}. As their name suggests, ON-OFF ganglion cells are able to respond to both ON and OFF edges moving in different directions with a broad range of speeds \cite{dhande2015contributions}. ON and OFF ganglion cells are sensitive to ON and OFF edges respectively. Meanwhile, as mentioned above, these direction-selective ganglion cells (DSGCs) are able to encode the size information of objects under the modulation of WACs.

\section{BIOINSPIRED RETINAL NEURAL NETWORK FOR MOTION DETECTION}
Based on the neural mechanisms of mammalian visual motion processing, a computational neural network model of primary visual motion perception is proposed in this section. Specifically, this model mainly contains four layers (Fig. \ref{fig:s02f01b}): (1) photoreceptor cells capture the change of luminance. (2) Bipolar cells encode the visual information into ON and OFF pathways, and perform fast and slow temporal filtering. (3) SACs perform direction-selective spatial filtering and WACs perform antagonistic center-surround spatial filtering. (4) RGCs discriminate motion patterns through computing the motion energy based on spatiotemporal filtering information.

\subsection{Photoreceptor cells}
The first layer consists of photoreceptor cells that receive grey-scale visual stimuli and capture the change of luminance. These cells are arranged as a 2D matrix that is the same size as the input image.
\begin{equation}
P(x,y,t)=I(x,y,t)-I(x,y,t-1)+\sum_i^{N_p}p_iP(x,y,t-i),
\end{equation}
where $P(x,y,t)$ represents the change of brightness corresponding to each pixel at time $t$, $x$ and $y$ are the abscissa and ordinate of the pixel respectively. $I(x,y,t)$ and $I(x,y,t-1)$ are the grey-scale value of two successive frames. Assume the change of luminance lasts for $N_p$ number of frames, but fast decays with coefficient $p_i$. Herein, we define $p_i=(1+e^{ui})^{-1}$, and choose $u=1$.

\subsection{Bipolar cells}
In the bipolar cells, the visual signal is split and processed separately in two parallel channels, an ON and an OFF channels. The ON channel responds to contrast increments and OFF channel responds to contrast decrements. As described in \cite{fu2018shaping}, a 'half-wave' rectifier can be used to simulate the ON-OFF split, where negative/positive inputs are fed to ON and OFF pathways, and negative inputs are inverted to positive in the OFF pathway.
\begin{equation}
\begin{split}
B_{+}(x,y,t)&=\frac{1}{2}\left[ |P(x,y,t)|+P(x,y,t)\right],\\
B_{-}(x,y,t)&=\frac{1}{2}\left[ |P(x,y,t)|-P(x,y,t)\right].
\end{split}
\end{equation}

Meanwhile, the bipolar cells receive some inhibitory signals from the horizontal cells and the amacrine cells. In this paper, difference of gaussians is used to model this process as a bandpass filter. The kernel function is the difference of two gaussians with different standard deviation
\begin{equation}
g_I(x,y;\sigma_1,\sigma_2)=\left(\frac{F}{\sqrt{2\pi}\sigma_1}e^{\frac{-(x^2+y^2)}{2\sigma_1^2}}-\frac{F}{\sqrt{2\pi}\sigma_2}e^{\frac{-(x^2+y^2)}{2\sigma_2^2}}\right),
\end{equation}
where $F$ is the gain coefficient, $\sigma_1$ and $\sigma_2$ are the standard deviation of two gaussians. The activations of the bipolar cells can be computed through convolving with this kernel function.
\begin{equation}
\begin{split}
B_{+}^0(x,y,t)&=B_{+}(x,y,t)\otimes g_I,\\
B_{-}^0(x,y,t)&=B_{-}(x,y,t)\otimes g_I,
\end{split}
\end{equation}
where $B_{+}^0(x,y,t)$ and $B_{-}^0(x,y,t)$ are the neural activity of ON and OFF bipolar cell in the first layer respectively. $\otimes$ represents the convolution operation.

In order to compute the neural activities with different response speed, a new temporal filtering method is proposed based on cascades of leaky integrators. The structure of this filter consists of n-level cascade of leaky integrators, denoted as $\mathcal{L}(K,n,z_0)$. $K$ is the gain of filter, $n$ is the number of layers, and $z_0$ is the input signal. Specifically, $n$th section of leaky integrators obeys the following dynamic equation.
\begin{equation}\label{eqns3h3}
\tau \frac{\mathrm{d}z_n}{\mathrm{d}t}=-Az_n+Cz_{n-1},
\end{equation}
where $\tau$ is a time constant of integration, $A$ is a decay coefficient, and $C$ is a coefficient of transmission. Finally, the output of temporal filtering is defined as the difference between two layers of neurons.
\begin{equation}\label{eqns3h4}
\mathcal{L}(K,n,z_0)=K(z_n-z_{n+m}),
\end{equation}
where $m>0$. In this paper, we choose $m=1$.

Then, the fast and slow responses of ON and OFF bipolar cells are computed respectively as follows
\begin{equation}
\begin{split}
B_{+}^f(x,y,t)&=\mathcal{L}(K, n_f, B_{+}^0),\\
B_{+}^s(x,y,t)&=\mathcal{L}(K, n_s, B_{+}^0),\\
B_{-}^f(x,y,t)&=\mathcal{L}(K, n_f, B_{-}^0),\\
B_{-}^s(x,y,t)&=\mathcal{L}(K, n_s, B_{-}^0),
\end{split}
\end{equation}
where $n_f<n_s$.

\subsubsection{Analysis of the temporal filtering} The response of this model to an instantaneous unit impulse can be calculated simply. For the dynamic Eqn.\ref{eqns3h3}, if $p$ is written for $\mathrm{d}/\mathrm{d}t$, the equation relating output $z_n$ to input $z_0$ is 
\begin{equation}\label{eqns3h5}
\left(p+\frac{A}{\tau}\right)^nz_n=\left( \frac{C}{\tau} \right) z_0.
\end{equation}
When $z_0$ is unit impulse signal, the solution of above equation is
\begin{equation}\label{eqns3h6}
z_n=\left( \frac{C}{\tau} \right)^n\frac{t^{n-1}e^{\frac{-At}{\tau}}}{(n-1)!}.
\end{equation}
Since output of temporal filtering $\mathcal{L}$ is the difference between $n$th and $(n+m)$th layers of neurons, the unit impulse response of Eqn.\ref{eqns3h4} is
\begin{equation}
\begin{split}
\mathcal{L}&=K\left( \frac{C}{\tau} \right)^n\frac{t^{n-1}e^{\frac{-At}{\tau}}}{(n-1)!}-K\left( \frac{C}{\tau} \right)^{n+m}\frac{t^{n+m-1}e^{\frac{-At}{\tau}}}{(n+m-1)!}\\
&=K\left(\frac{C}{\tau}\right)^nt^{n-1}e^{\frac{-At}{\tau}}\left[ \frac{1}{(n-1)!}-\frac{\left( \frac{Ct}{\tau}\right)^m}{(n+m-1)!}\right].
\end{split}
\end{equation}
Let $a=\frac{A}{\tau}$, $b=\frac{C}{\tau}$, then we have 
\begin{equation}
\mathcal{L}=Ke^{-at}\left[ \frac{b^nt^{n-1}}{(n-1)!} - \frac{b^{n+m}t^{n+m-1}}{(n+m-1)!} \right].
\end{equation}
The derivative with respect to $t$ is 
\begin{equation}
\begin{split}
\frac{\mathrm{d}\mathcal{L}}{\mathrm{d}t}&=-Kae^{-at}\left[ \frac{b^nt^{n-1}}{(n-1)!} - \frac{b^{n+m}t^{n+m-1}}{(n+m-1)!} \right]\\
&\quad +Ke^{-at}\left[ \frac{b^nt^{n-2}}{(n-2)!}-\frac{b^{n+m}t^{n+m-2}}{(n+m-2)!} \right].
\end{split}
\end{equation}
Let $\frac{\mathrm{d}\mathcal{L}}{\mathrm{d}t}=0$, we have 
\begin{equation}
\begin{split}
\left[ \frac{b^nt^{n-2}}{(n-2)!}-\frac{b^{n+m}t^{n+m-2}}{(n+m-2)!} \right]
-a\left[ \frac{b^nt^{n-1}}{(n-1)!} - \frac{b^{n+m}t^{n+m-1}}{(n+m-1)!} \right]=0.
\end{split}
\end{equation}
Simplify this equation, there is 
\begin{equation}
\frac{n-1-at}{(n-1)!}-\frac{b^mt^m(n+m-1-at)}{(n+m-1)!}=0.
\end{equation}
Then we have
\begin{equation}
\begin{split}
&ab^m(n-1)!t^{m+1}-b^m(n+m-1)(n-1)!t^m\\
&-a(n+m-1)!t+(n-1)(n+m-1)!=0.
\end{split}
\end{equation}
Especially when $m=1$, we have
\begin{equation}
ab(n-1)!t^2-(a+b)n!t+(n-1)n!=0.
\end{equation}
Since 
\begin{equation}
\begin{split}
\Delta &=\left[ (a+b)n! \right]^2 -4ab(n-1)! \cdot (n-1)n!\\
&=(an!)^2+(bn!)^2+2ab(n!)^2-4ab(n-1)!\cdot (n-1) n! \\
&>(an!)^2+(bn!)^2+2ab(n!)^2-4ab(n!)^2\\
&=\left[ (a-b)n! \right]^2 \geq 0,
\end{split}
\end{equation}
there are two real roots.
\begin{equation}
t_1=\frac{(a+b)n!-\sqrt{\Delta}}{2ab(n-1)!},
\end{equation}
\begin{equation}
t_2=\frac{(a+b)n!+\sqrt{\Delta}}{2ab(n-1)!}.
\end{equation}
The activity reaches a maximum at time $t_1$, and reaches a minimum at time $t_2$. When $A>C$ ($a>b$), then $t_1<\frac{n}{a}$ and $t_2>\frac{n}{b}$. Otherwise when $A<C$ ($a<b$), then $t_1<\frac{n}{b}$ and $t_2>\frac{n}{a}$.

\begin{figure}[!htbp]
    \centering
    \subfigure[]{\label{fig:s03f02a}\includegraphics[width=0.4\textwidth]{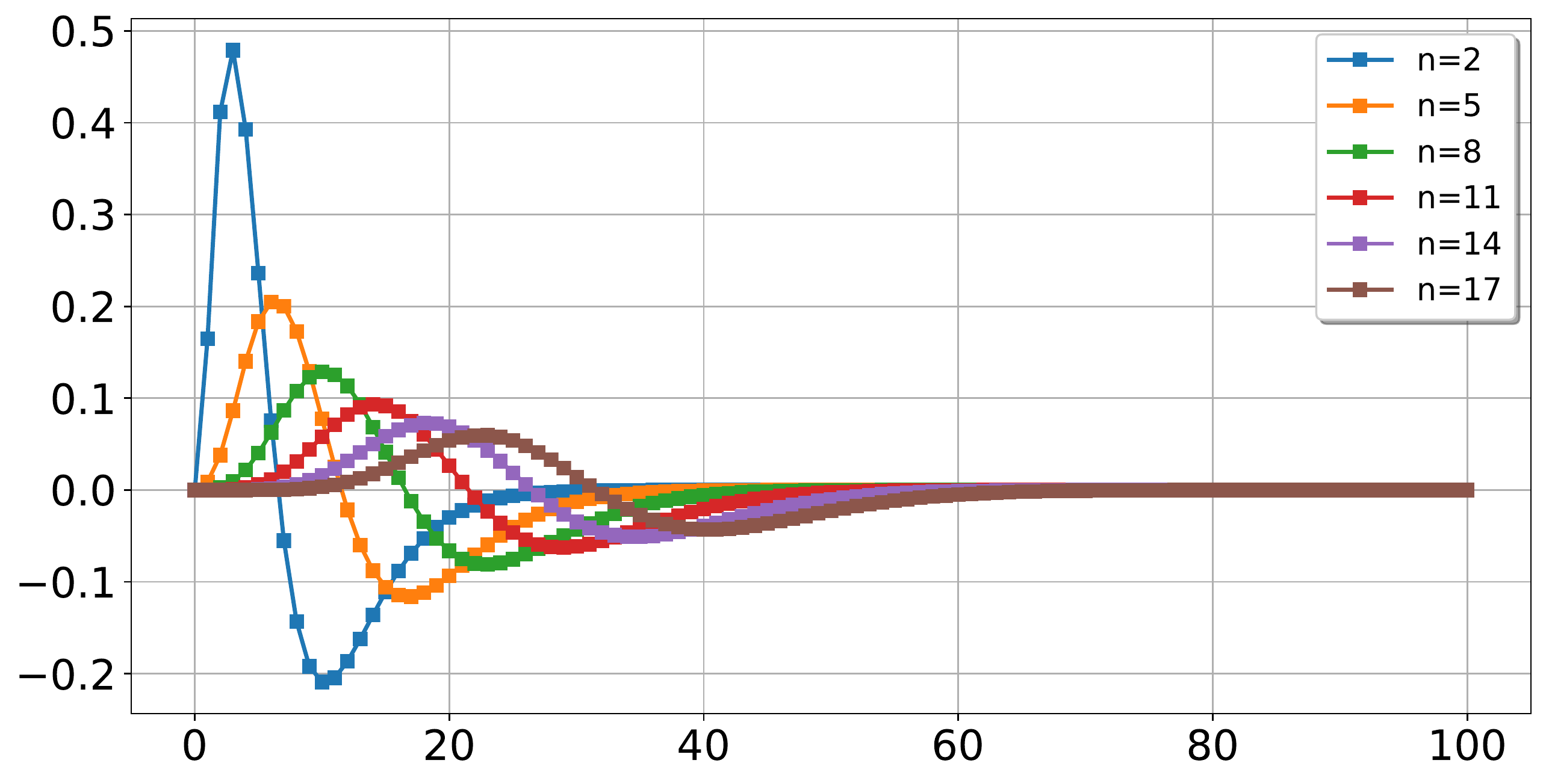}}\\
    \subfigure[]{\label{fig:s03f02b}\includegraphics[width=0.4\textwidth]{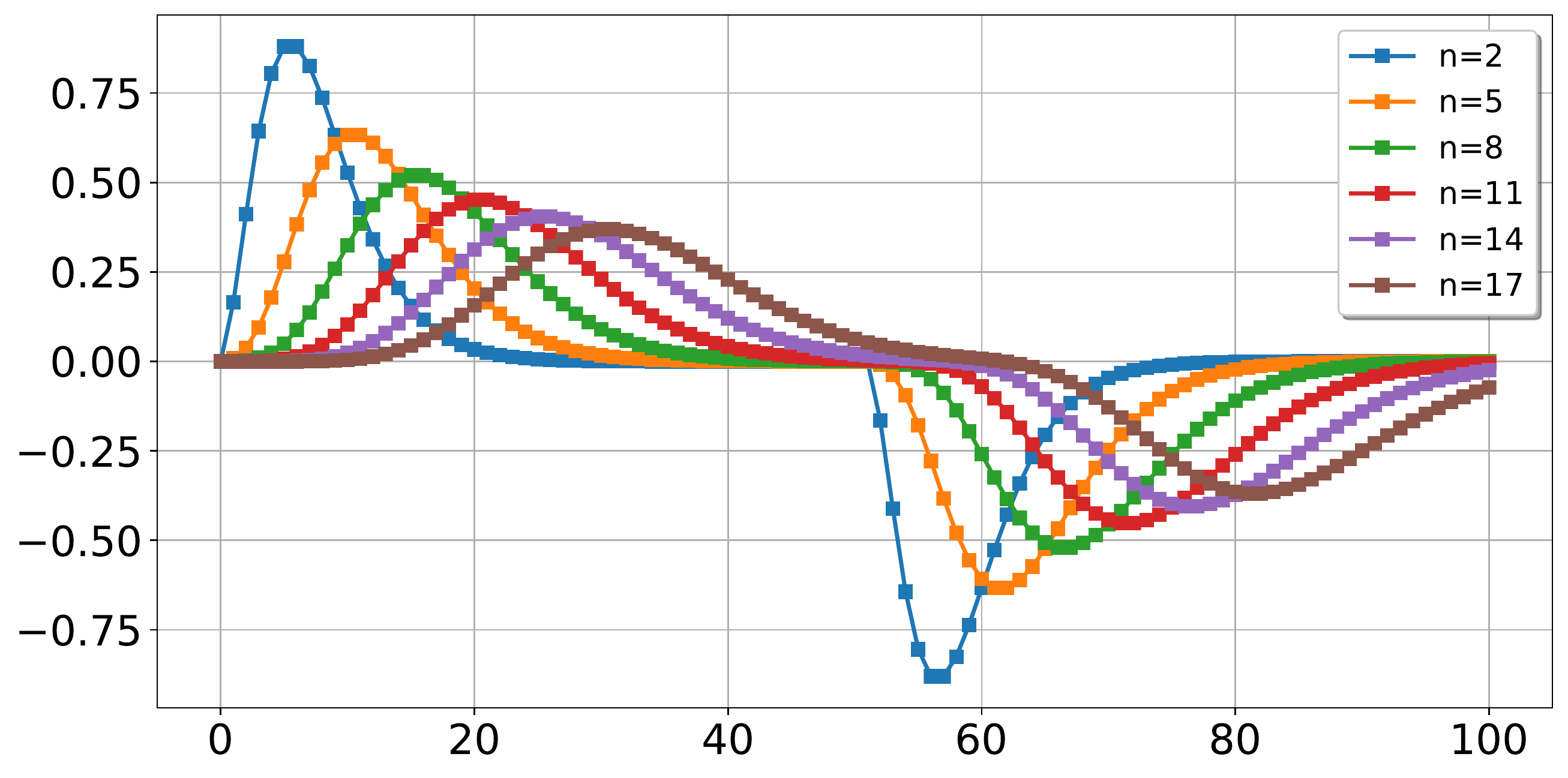}}
    \caption{The response characteristics of the neurodynamics-based temporal filter. (a) The unit impulse response of temporal filter at different $n$. (b) The step response of temporal filter at different $n$.}
    \label{fig:s03f02}
\end{figure}

We conduct a simple simulation to observe the characteristics of this neurodynamics-based temporal filter. The total simulation time is 100 steps, and each step is set as $\Delta t =0.05s$. The gain of filter $K=5$, the decay coefficient $A=60$, the coefficient of transmission $C=60$, and $\tau=8$. The unit impulse response and step response of temporal filter at different $n$ are shown in Fig. \ref{fig:s03f02a} and Fig. \ref{fig:s03f02b} respectively. Obversely, the response speed decreases with the increase of $n$. Meanwhile, the response amplitude is also related to the number of cascades $n$. As $n$ increases, the response gradually weakens.



Actually, the above response is similar to the temporal impulse responses described in \cite{adelson1985spatiotemporal}, where the temporal filter in the model is also biphasic. The filtering result is computed offline through convolving or correlating the signal history with the following kernel function.
\begin{equation}
f(t)=\frac{(kt)^n}{n!}e^{-kt}-\frac{(kt)^{n+2}}{(n+2)!}e^{-kt}.
\end{equation}
Although it can serve as plausible approximations to psychophysical data, it is not biologically 
plausible for the neural computation process. In addition, this method needs convolution or correlation along the time dimension, which usually suffers from typical phase delay or phase shift problem such that real-time performance is affected. While the proposed filter performs the temporal filtering only through reading out the activity of Leaky neurons directly. It is easy to reduce phase delay and shift by selecting the superficial neural outputs for improving response speed.

\subsection{Amacrine cells}
\subsubsection{Direction-selective Spatial Filtering} 
\begin{figure}[!htbp]
    \centering
    \subfigure[]{\label{fig:s03f03a}\includegraphics[width=0.4\textwidth]{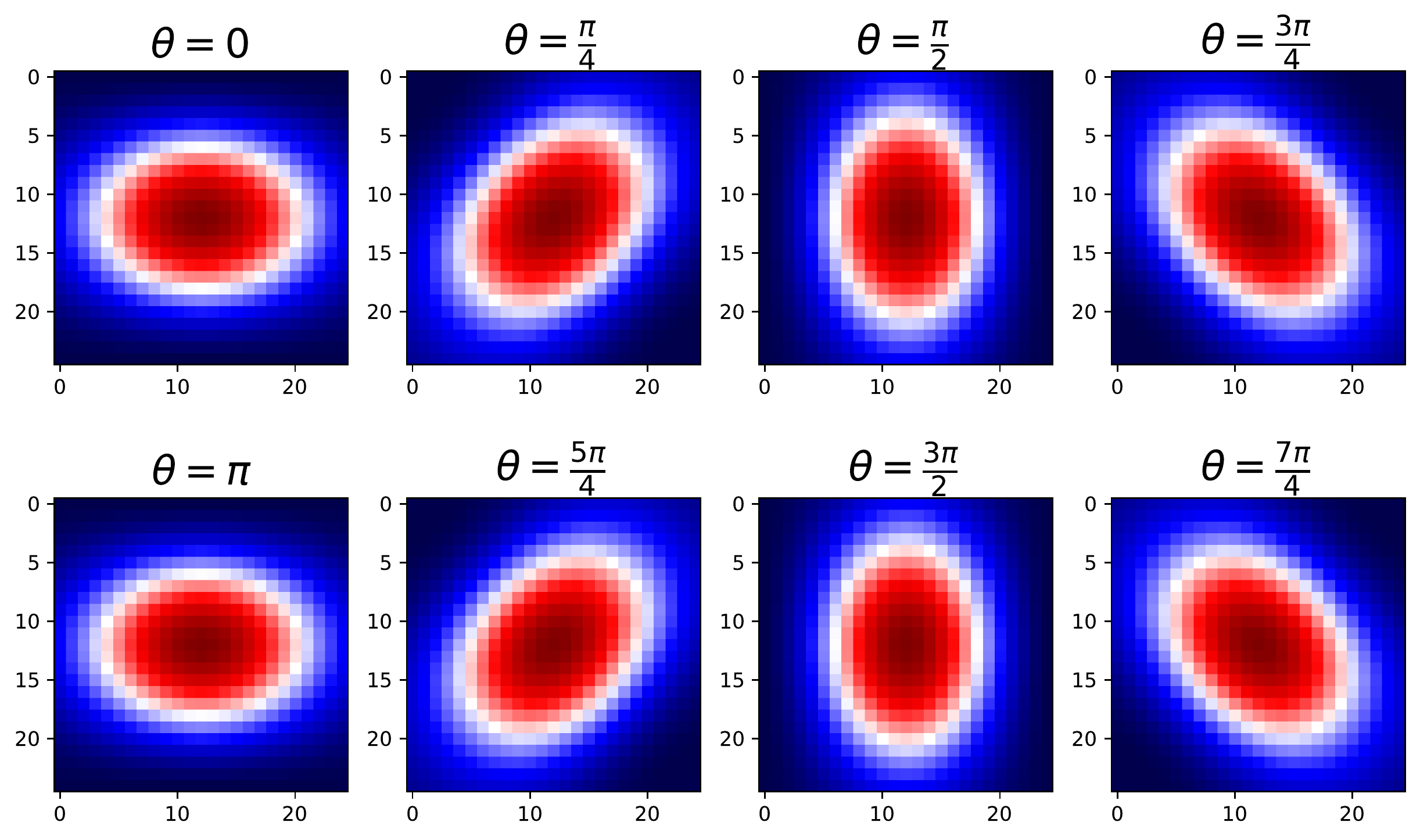}}\\
    \subfigure[]{\label{fig:s03f03b}\includegraphics[width=0.4\textwidth]{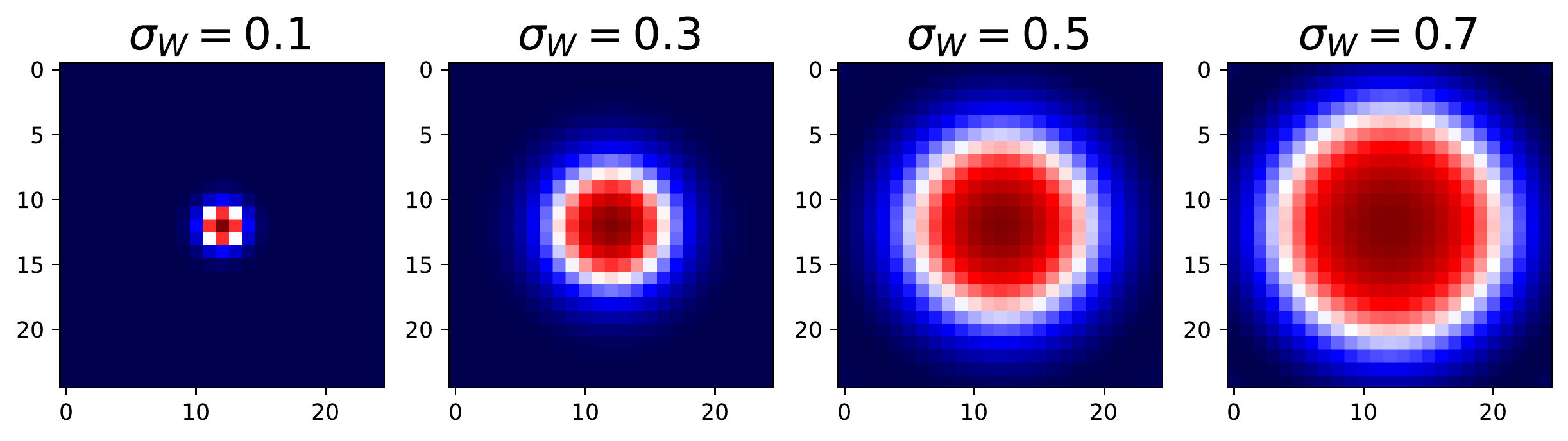}}
    \caption{The shape of spatial filters with different parameters. (a) The shape of direction-selective Gabor filters at different orientations. (b) The shape of antagonistic center-surround Gabor filters at different standard deviations.}
    \label{fig:s03f03}
\end{figure}

SACs perform direction-selective spatial filtering. Frequency and orientation representations of Gabor filters are demonstrated to be similar to those of the human visual system. Thus, Gabor filters are used to perform spatial filtering of fast and slow bipolar neurons. The kernel function is
\begin{equation}
g_S(x,y;\lambda_S,\theta,\sigma_S)=\exp\left(-\frac{x^{\prime 2}+y^{\prime 2}}{2\sigma_S^2}\right)\cos\left(\frac{2\pi x^{\prime}}{\lambda_S}+\psi\right),
\end{equation}
where $x'=x\cos(\theta)+y\sin(\theta)$, $y'=-x\sin(\theta)+y\cos(\theta)$. $\lambda_S$ is wavelength of the sinusoidal factor, which is usually measured in pixels. $\theta$ represents the orientation of the Gabor function. $\psi$ is the phase offset. $\sigma_S$ is the standard deviation of the Gaussian envelope. The shape of Gabor kernel function at different orientations is shown in Fig. \ref{fig:s03f03a}, where $\lambda_S=2, \sigma_S=3, \psi=0$.

Herein, two gabor filters are used to process the visual information from the bipolar cells. The phase offset of the first filter $g_1$ is $\psi_1=0$, and the one of the second filter $g_2$ is $\psi_2=\frac{\pi}{2}$. Then, for each orientation $\theta$, the neural activities of the starburst amacrine cells are calculated in the following way.
\begin{equation}
\begin{split}
	SA_{+}^1(x,y,t)&=B_{+}^s(x,y,t)\otimes g_1,\\
	SB_{+}^1(x,y,t)&=B_{+}^s(x,y,t)\otimes g_2,\\
	SA_{+}^2(x,y,t)&=B_{+}^f(x,y,t)\otimes g_1,\\
	SB_{+}^2(x,y,t)&=B_{+}^f(x,y,t)\otimes g_2,\\
	SA_{-}^1(x,y,t)&=B_{-}^s(x,y,t)\otimes g_1,\\
	SB_{-}^1(x,y,t)&=B_{-}^s(x,y,t)\otimes g_2,\\
	SA_{-}^2(x,y,t)&=B_{-}^f(x,y,t)\otimes g_1,\\
	SB_{-}^2(x,y,t)&=B_{-}^f(x,y,t)\otimes g_2.
\end{split}
\end{equation}
where $\otimes$ represents the convolution operation. Eight kinds of SACs are defined here for computing the motion energy in the ganglion cells.

\subsubsection{Antagonistic Center-surround Spatial Filtering} WACs mainly perform antagonistic center-surround spatial filtering to control size selectivity. Herein, a new kernel function is defined following the formation of Gabor filters. It can excite the center and inhibit the surrounding without direction selectivity. 
\begin{equation}
g_W(x,y;\lambda_W,\sigma_W)=2\exp\left(-\frac{x^2+y^2}{2\sigma_W^2}\right)\cos\left(\frac{2\pi x}{\lambda_W}\right)-1,
\end{equation}
where $\lambda_W$ is wavelength of the sinusoidal factor, $\sigma_W$ is the standard deviation of the Gaussian envelope. The kernel size is denoted as $M_W\times M_W$. The balance of excitation/inhibition is mainly regulated by the kernel size $M_W$ and the standard deviation $\sigma_W$, which plays a dominant role in controlling size selectivity. The shape of this kernel function at different $\sigma_W$ is displayed in Fig. \ref{fig:s03f03b}, where $M_W=25$.

WACs contact bipolar cell terminals and generate mediated signals to control ganglion cell responses. Specifically, the ON and OFF mediated signals are obtained as follows.
\begin{equation}
\begin{split}
	WA_{+}(x,y,t)&=\left[ \left[ B_{+}^s(x,y,t)\right]^{+}\otimes g_W \right]^{+},\\
	WA_{-}(x,y,t)&=\left[ \left[ B_{-}^s(x,y,t)\right]^{+}\otimes g_W \right]^{+}.
\end{split}
\end{equation}
where $[\cdot]^{+}=\max(0,\cdot)$ represents a rectified linear unit. This process can boost center and inhibit surrounding response based on activations of bipolar cells.

\subsection{Retinal ganglion cells}
\subsubsection{Motion Direction Estimation} RGCs can use the spatiotemporal filtering information to compute the motion energy of different motion patterns. In the motion direction sensing task, ganglion cells fuse the spatiotemporal filtering results of different phases and different speeds to calculate the motion energy of each pixel. According to the motion energy in each direction, then, the direction can be discriminated. Following the Hassenstein-Reichardt model \cite{werner1961autocorrelation} and Adelson-Bergen model \cite{adelson1985spatiotemporal}, the motion energy at each filtering orientation $\theta$ can be computed as follows.
\begin{equation}
\begin{split}
E^{\theta}_{+}(x,y,t)&=SA_{+}^1(x,y,t)\cdot SB_{+}^2(x,y,t)\\
&\quad -SA_{+}^2(x,y,t)\cdot SB_{+}^1(x,y,t),\\
E^{\theta}_{-}(x,y,t)&=SA_{-}^1(x,y,t)\cdot SB_{-}^2(x,y,t)\\
&\quad-SA_{-}^2(x,y,t)\cdot SB_{-}^1(x,y,t).
\end{split}
\end{equation}

In addition to the separation of ON and OFF pathways, there are several types of retinal ganglion cells that recombine ON and OFF signals from SACs, such as ON-OFF direction-selective retinal ganglion cells. This type of cells can respond to both contrast increments and decrements along a particular direction. Herein, ON-OFF motion energy is defined in a weighted summation of ON and OFF pathways.
\begin{equation}
E^{\theta}(x,y,t)=w_{+}E_{+}^{\theta}(x,y,t)+w_{-}E_{-}^{\theta}(x,y,t),
\end{equation}
where $w_{+}$ and $w_{-}$ are weighting coefficients of the recombination of ON and OFF signals. The motion direction can be estimated with only two mutually perpendicular spatial filtering processes, which are the motion energy at $0$ and $\frac{\pi}{2}$ respectively.
\begin{equation}
\hat{\varphi} = \arctan2(E^{\frac{\pi}{2}}(x,y,t), E^{0}(x,y,t)).
\end{equation}

\subsubsection{Size-mediated Motion Detection with Moving Background Suppresion} WAC-mediated responses of RGCs are computed by combining the maximal motion energy and the mediated signals. Assume $\mathbf{E}^{\Theta}(x,y,t)$ is the set of neural outputs at all filtering orientations $\Theta$, $\Theta=\{0,\frac{\pi}{4},\frac{\pi}{2},\frac{3\pi}{4},\pi,\frac{5\pi}{4},\frac{3\pi}{2},\frac{7\pi}{4}\}$. Then activities of size-selective ganglion cells are determined in the following way.
\begin{equation}
\begin{split}
V_{+}(x,y,t)&=\max_{\theta}\mathbf{E}^{\Theta}_{+}(x,y,t)\cdot WA_{+}(x,y,t),\\
V_{-}(x,y,t)&=\max_{\theta}\mathbf{E}^{\Theta}_{-}(x,y,t)\cdot WA_{-}(x,y,t).
\end{split}
\end{equation}
In the same way, the ON and OFF pathways are combined through the weighted summation.  
\begin{equation}
V(x,y,t)=\left [ w_{+}V_{+}(x,y,t)+w_{-}V_{-}(x,y,t)\right ]^+,
\end{equation}
It is worth noting that the activation of RGC $V(x,y,t)$ synthesizes the information of motion energy and size of the moving object, which is jointly mediated by contrast, motion velocity, motion direction and target size. It is an ideal object for performing small-target motion detection.

Specifically, there are two steps to execute this procedure. Firstly, $Norm(V)$ are used to extract regions of interest, where $Norm(\cdot)$ is a Min-Max normalization function. For a given detection threshold $\gamma$, if $Norm(V(x,y,t))>\gamma$, the location $(x,y)$ is added to the regions of interest. Then the DBSCAN \cite{ester1996density,schubert2017dbscan} algorithm is used to discover the clusters of excitatory neural populations. The clustering center $(X,Y)$ is regarded as the position of the candidate target. The motion direction can be estimated by a population coded algorithm. The maximal motion energy of each point is decomposed into two parts along the $x$ and $y$ axis.
\begin{equation}
\begin{split}
E_{x}(x,y,t)&=\max_{\theta} \mathbf{E}^{\Theta}(x,y,t)\cos (\hat{\varphi}(x,y,t)),\\
E_{y}(x,y,t)&=\max_{\theta} \mathbf{E}^{\Theta}(x,y,t)\sin (\hat{\varphi}(x,y,t)).
\end{split}
\end{equation}
Then the maximal motion energy of each cluster can be computed via average method.
\begin{equation}
\begin{split}
E_X(x,y,t)&=\frac{1}{N_{\mathcal{T}}}\sum_{(x,y)\in \mathcal{T}} E_x(x,y,t),\\
E_Y(x,y,t)&=\frac{1}{N_{\mathcal{T}}}\sum_{(x,y)\in \mathcal{T}} E_y(x,y,t).
\end{split}
\end{equation}
where $\mathcal{T}$ is the point set of a candidate target, $N_{\mathcal{T}}$ is the number of points. The motion direction and energy of the moving target can be estimated as follows.
\begin{equation}
\begin{split}
\varPhi(X,Y,t)&=\arctan2(E_X(x,y,t), E_Y(x,y,t)),\\
E(X,Y,t)&=\sqrt{E_X(x,y,t)^2+E_Y(x,y,t)^2}.
\end{split}
\end{equation}

\begin{figure}[!htbp]
    \centering
    \includegraphics[width=0.5\textwidth]{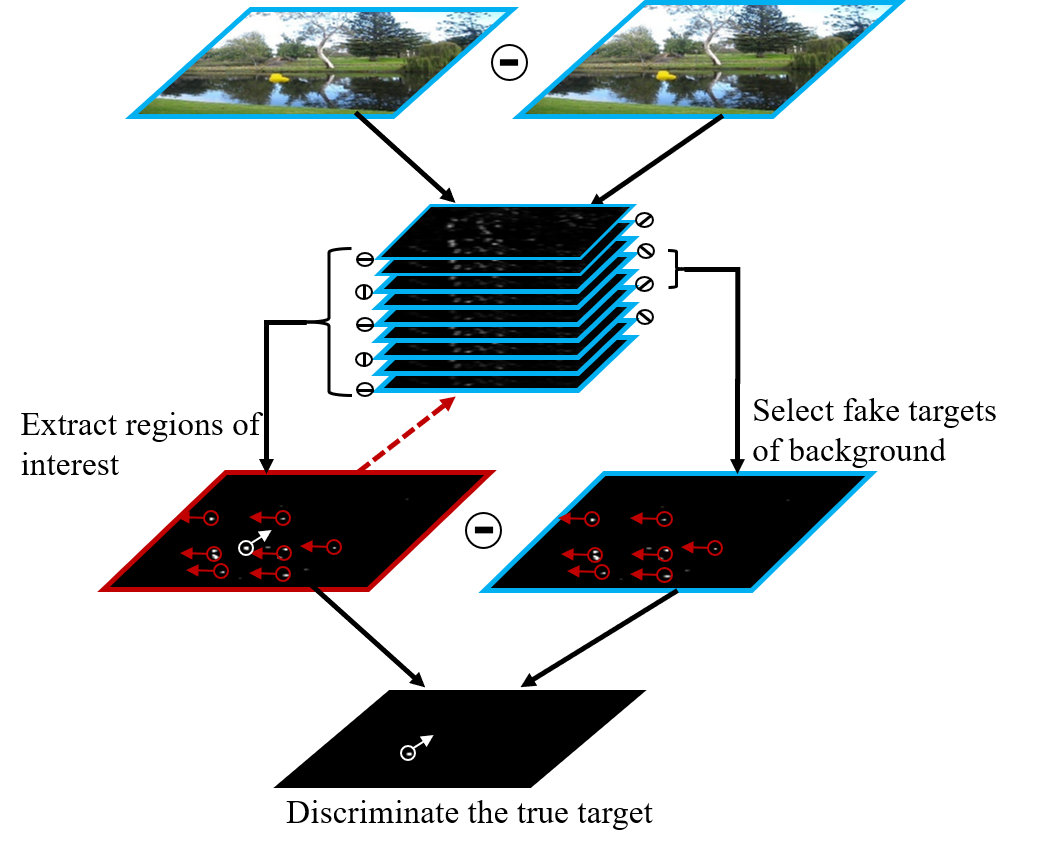}
    \caption{The diagram of directionally selective inhibition.}
    \label{fig:s03f05}
\end{figure}

In the real world, a moving background brings huge disturbance to motion detection since many of the features are similar to small moving targets in this situation. As one of key characteristics, these fake targets usually moves in the same direction when the camera shakes. In order to decrease this disturbance, we propose a background suppression method based on the algorithm of directionally selective inhibition, as shown in Fig. \ref{fig:s03f05}. The motion direction ($0-2\pi$) is equally divided into eight intervals. The number of targets is then counted in each interval. If the number of targets is greater than a threshold $N_{th}$ in the internal $[a, b]$, $a$ and $b$ are added to the set of inhibitory direction $\Theta_I$. Then the activities of RGCs with directionally selective inhibition become the following form.
\begin{equation}
\begin{split}
V'_{+}(x,y,t)&=V_{+}(x,y,t) - \max_{\theta_I}\mathbf{E}^{\Theta_{I}}_{+}(x,y,t)\cdot WA_{+}(x,y,t),\\
V'_{-}(x,y,t)&=V_{-}(x,y,t) -\max_{\theta_I}\mathbf{E}^{\Theta_{I}}_{-}(x,y,t)\cdot WA_{-}(x,y,t).
\end{split}
\end{equation}
Finally, $Norm(V')$ is used to determine the target location. In the same way, the location $(x,y)$ is added to the region of interest if $Norm(V'(x,y,t))>\gamma$. Then the DBSCAN algorithm is used to determine the position of targets.

Algorithm \ref{alg_1} presents the pseudocode for the proposed model.
\IncMargin{1em}
\begin{algorithm}[!htbp]
\label{alg_1}
\caption{Bioinspired Retinal Neural Network.}
\SetKwFunction{ComputePhotoreceptor}{ComputePhotoreceptor}
\SetKwFunction{ComputeFirstBipolar}{ComputeFirstBipolar}
\SetKwFunction{TemporalFiltering}{TemporalFiltering}
\SetKwFunction{DSFiltering}{DSFiltering}
\SetKwFunction{CSFiltering}{CSFiltering}
\SetKwFunction{MotionEnergy}{MotionEnergy}
\SetKwFunction{DirectionEstimation}{DirectionEstimation}
\SetKwFunction{ComputeGanglion}{ComputeGanglion}
\SetKwFunction{DirectionalInh}{DirectionalInh}
\SetKwFunction{ContrastEnhancing}{ContrastEnhancing}
\SetKwFunction{VelocityEstimation}{VelocityEstimation}
\SetKwInOut{Input}{input}\SetKwInOut{Output}{output}
\KwIn{The grayscale image of current frame.}
\KwOut{The estimated position, direction $\hat{\varphi}$ and velocity $\hat{v}$ of small moving targets.}
\BlankLine
Initialize the parameters of the neural network, and record the initial image $I_1$\;
\For{$t \leftarrow 2$ \KwTo $T$}{
Store the previous image $I_{t-1}$, and get the current image $I_t$\;
$P_t\leftarrow $ \ComputePhotoreceptor{$I_t$,$I_{t-1}$}\;
$B_{+}^0,B_{-}^0\leftarrow$ \ComputeFirstBipolar{$P_t$,$g_I$}\;
\tcp{\emph{Perform temporal filtering}}
$B_{+}^f,B_{+}^s\leftarrow$ \TemporalFiltering{$B_{+}^0$}\;
$B_{-}^f,B_{-}^s\leftarrow$ \TemporalFiltering{$B_{-}^0$}\;
\For{$\theta \leftarrow 0$ \KwTo $2\pi$}{
\tcp{\emph{Perform direction-selective spatial filtering}}
$SA_{+}^1\leftarrow$ \DSFiltering{$B_{+}^s,g_1,\theta$}\;
$SB_{+}^1\leftarrow$ \DSFiltering{$B_{+}^s,g_2,\theta$}\;
$SA_{+}^2\leftarrow$ \DSFiltering{$B_{+}^f,g_1,\theta$}\;
$SB_{+}^2\leftarrow$ \DSFiltering{$B_{+}^f,g_2,\theta$}\;
$SA_{-}^1\leftarrow$ \DSFiltering{$B_{-}^s,g_1,\theta$}\;
$SB_{-}^1\leftarrow$ \DSFiltering{$B_{-}^s,g_2,\theta$}\;
$SA_{-}^2\leftarrow$ \DSFiltering{$B_{-}^f,g_1,\theta$}\;
$SB_{-}^2\leftarrow$ \DSFiltering{$B_{-}^f,g_2,\theta$}\;
\tcp{\emph{Perform antagonistic center-surround spatial filtering}}
$WA_{+}\leftarrow$ \CSFiltering{$B_{+}^s,g_W$}\;
$WA_{-}\leftarrow$ \CSFiltering{$B_{-}^s,g_W$}\;
\tcp{\emph{Compute motion energy}}
$E_{+}^{\theta},E_{-}^{\theta},E^{\theta}\leftarrow$ \MotionEnergy{$SA_{+}^1,SB_{+}^1,SA_{+}^2,SB_{+}^2$}\;
$\theta\leftarrow \theta + \frac{\pi}{4}$ 
}
$\hat{\varphi}\leftarrow$ \DirectionEstimation{$E^0,E^{\frac{\pi}{2}}$}\;
\tcp{Compute size-mediated ganglion activations.}
$V_{+}\leftarrow$ \ComputeGanglion{$\mathbf{E}_{+}^{\Theta},WA_{+}$}\;
$V_{-}\leftarrow$ \ComputeGanglion{$\mathbf{E}_{-}^{\Theta},WA_{-}$}\;
Obtain the set of inhibitory direction $\Theta_I$\;
\tcp{Perform directionally selective inhibition}
$V'_{+}\leftarrow$ \DirectionalInh{$\mathbf{E}_{+}^{\Theta},\mathbf{E}_{+}^{\Theta_I},WA_{+}$}\;
$V'_{-}\leftarrow$ \DirectionalInh{$\mathbf{E}_{-}^{\Theta},\mathbf{E}_{-}^{\Theta_I},WA_{-}$}\;
Obtain the positions of small targets through using the DBSCAN algorithm\;
}
\end{algorithm}
\DecMargin{1em}

\section{EXPERIMENTAL RESULTS AND DISCUSSION}
In this section, three groups of experiments are conducted to verify the the accuracy and effectiveness of the proposed algorithm. The first one is to extract the motion information of moving target in white background, which is used to test the basic properties of the proposed model. The second one is the small-target motion detection experiment in a synthetic scenario. The final one is to test the detection ability of small moving targets in the real cluttered natural environments based on the STNS dataset.

\subsection{Basic Properties of the Proposed Model}
\begin{figure*}[!htbp]
    \centering
    \subfigure[]{\label{fig:s04f01a}\includegraphics[width=0.16\textwidth]{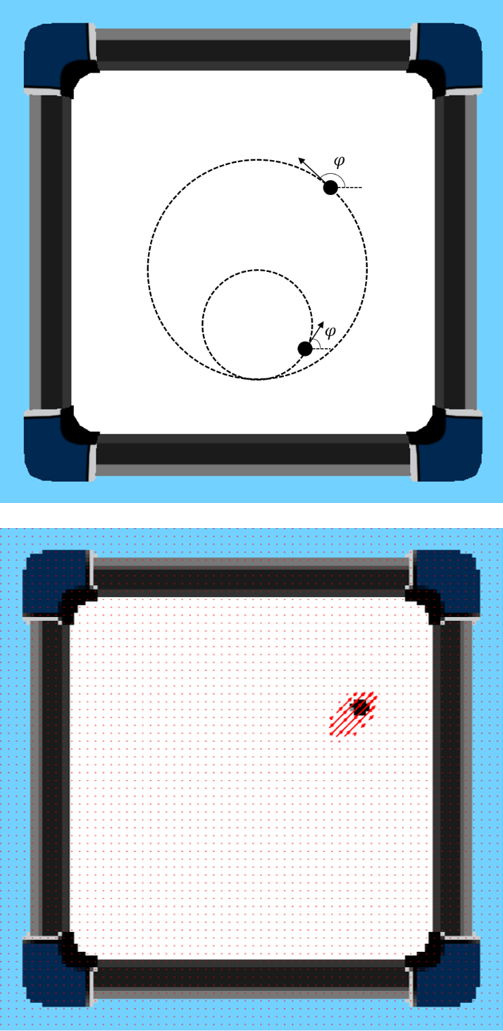}}%
    ~\hspace{15pt}
    \subfigure[]{\label{fig:s04f01b}\includegraphics[width=0.32\textwidth]{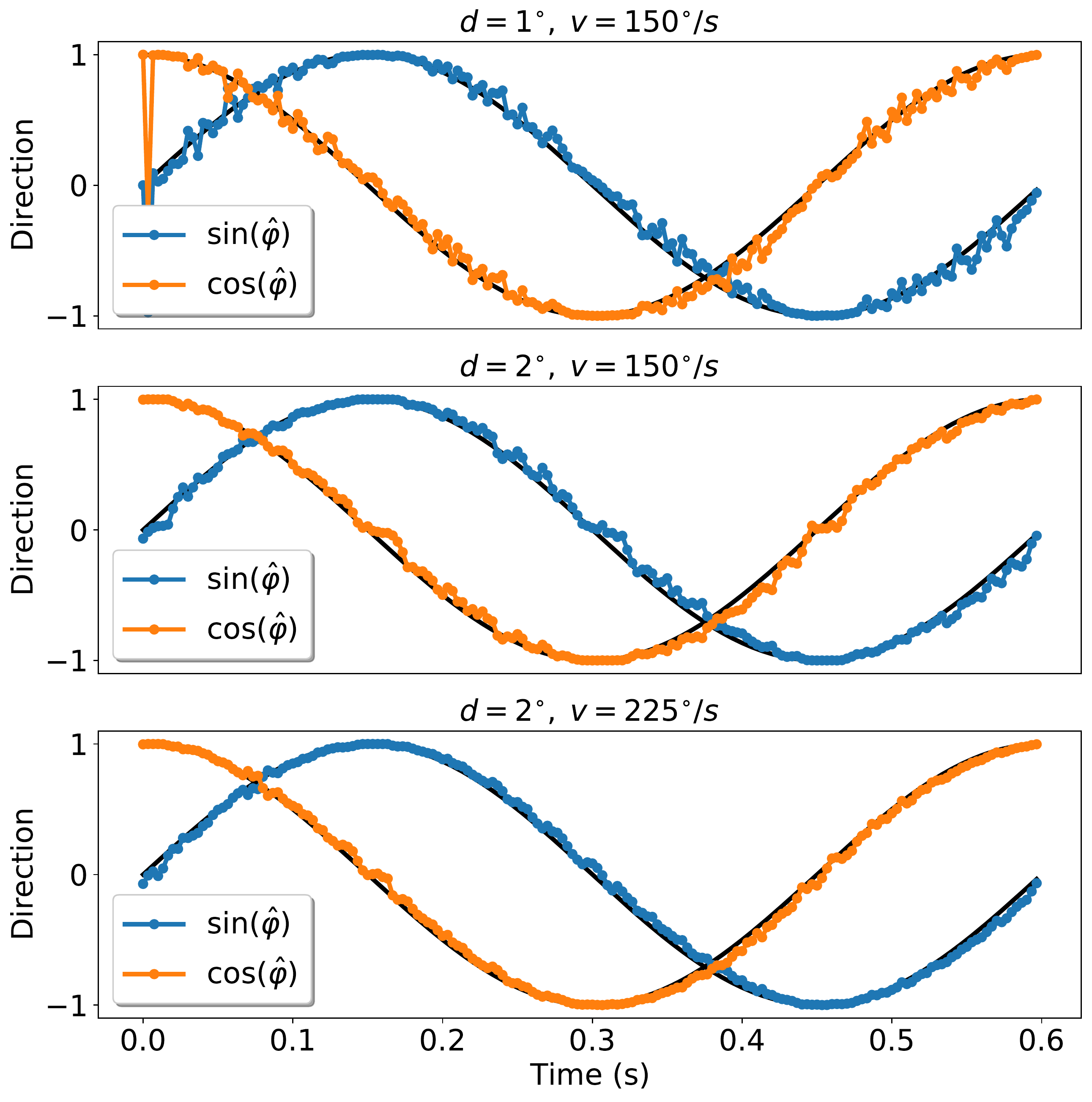}}
    ~\hspace{15pt}
    \subfigure[]{\label{fig:s04f01c}\includegraphics[width=0.32\textwidth]{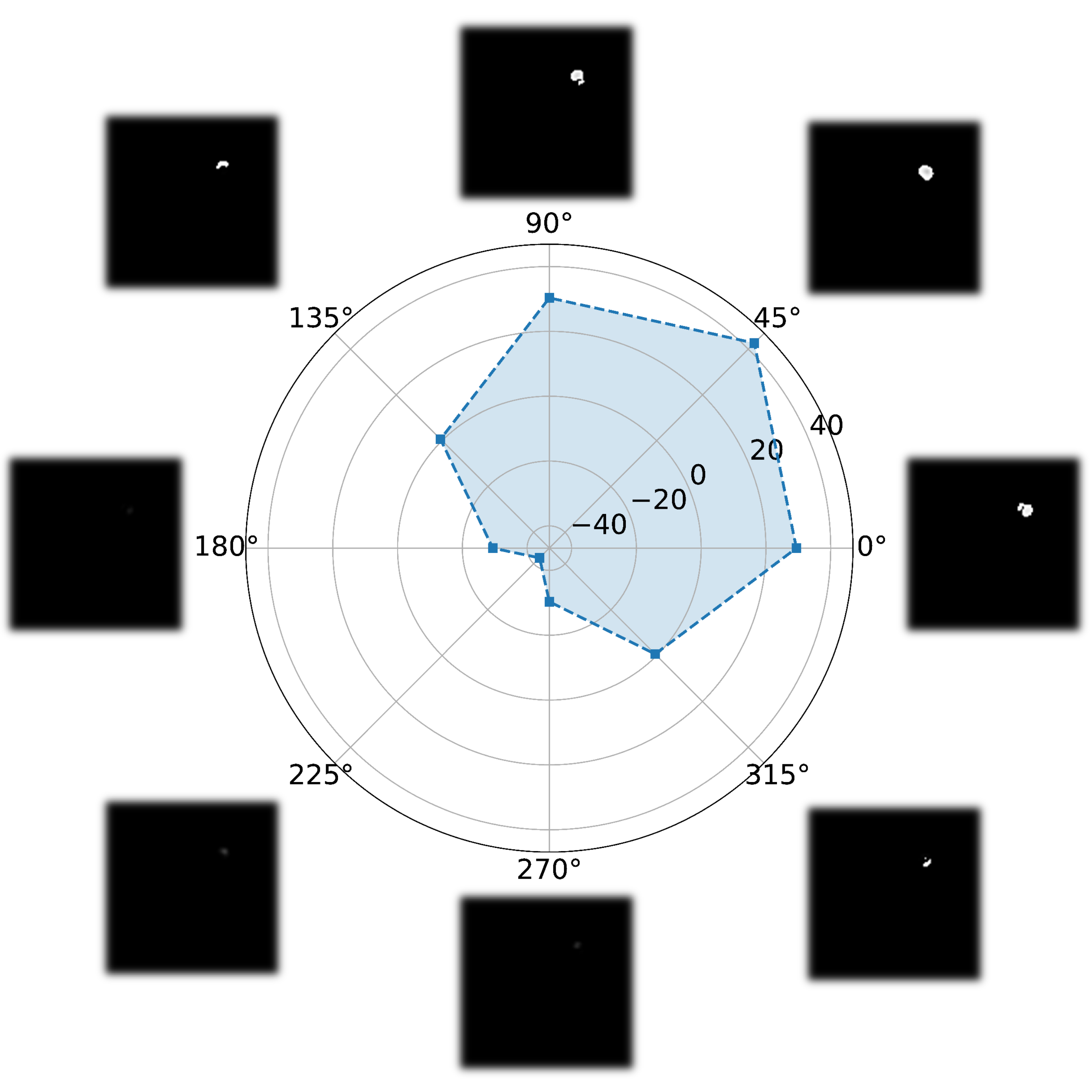}}
    \caption{Direction selectivity of the proposed BRNN. (a) Two cases of small-target direction perception. (b) Direction estimation at different velocities and  target sizes (circular motion). (c) The polar diagrams of motion energy at different filtering orientations ($45^{\circ}$ motion).}
    \label{fig:s04f01}
\end{figure*}

Based on the open source toolkit Psychlab developed by DeepMind, we firstly build a simple scene of small moving target, as shown in Fig. \ref{fig:s04f01a}. To evaluate the performance of direction estimation, we let a small target move along a circular orbit. Then a 45-degree motion is selected to analyse the property of our proposed model.

\subsubsection{Direction Selectivity}
At each time step, the input image is $128\times128$ gray-scale map. The number of total execution time steps is 180, and the frame rate is 300Hz. For the temporal filtering in the bipolar cells, we choose the decay coefficient $A=60$, the time constant of integration $\tau=5$, $\Delta t=0.05s$, the gain $K=5$, $n_f=2$ for the fast response, and $n_s=4$ for the slow response. For the spatial filtering in the SACs, the size of direction-selective Gabor kernel function is $5\times5$, the wavelength $\lambda_S=4$, the standard deviation of the Gaussian envelope $\sigma_S=0.3$. The field of view (FOV) is assumed as FOV$\sim 32^{\circ}$. By default, the target diameter is $1^{\circ}$ and the motion velocity is set to $150^{\circ}/s$. Different target sizes and velocities are chosen to evaluate the performance of direction estimation. The results show the proposed model has a good selectivity of motion direction, as displayed in Fig. \ref{fig:s04f01b}. The sine and cosine of the estimated direction can track the real motion direction accurately. Meanwhile, the estimation accuracy is higher when the size and velocity are relatively large. In fact, the closer the size and velocity are to the optimal value, the higher the estimation accuracy is.

In the case of 45-degree motion, we record the sum of motion energy defined as follows.
\begin{equation}
E(\theta,t)=\sum_{(x,y)\in \mathcal{T}} E_{+}^{\theta}(x,y,t)+\sum_{(x,y)\in \mathcal{T}} E_{-}^{\theta}(x,y,t),
\end{equation}
where $E_{+}^{\theta}(x,y,t)$ and $E_{-}^{\theta}(x,y,t)$ represent the motion energy at the orientation $\theta$ respectively. $\mathcal{T}$ represents the point set of the detected target. The Fig. \ref{fig:s04f01c} shows the polar diagram of the sum of motion energy at different orientations. The value at $\theta=45^{\circ}$ is largest, which demonstrates a good ability of determining the motion direction. Meanwhile, the activities of neurons whose filtering orientation ($135^{\circ}$ and $315^{\circ}$) is perpendicular to the real motion direction are staying at zero all the time. If the angle between the filtering orientation and the motion direction is acute, the activations of neurons are positive. On the contrary, if the angle is obtuse, the activations are negative. For the neural activities, the response is strong in the situation where the filtering orientation is consistent with the motion direction, which can reflects the correlation between the filtering orientation and the motion direction well. The closer the filtering orientation and the motion direction is, the higher the intensity is. It results from two reasons. Firstly, the spatial filtering can select the motion patterns in or against the real motion direction. But the order of signal change in these two motion patterns is different, the temporal filtering can inhibit the neural activities in the opposite-direction motion.

\subsubsection{Contrast Sensitivity}
Contrast sensitivity is one of the important attributes of primary motion perception. Generally, high-contrast visual object is much easier to be found. The visual response increases with the increase of contrast, which is called contrast sensitivity. Herein, Weber contrast is used to test the contrast sensitivity, which is defined as $c=\frac{|\bar{I}_t-\bar{I}_b|}{255}$. $\bar{I}_t$ is the average grey-scale value of moving target, and $\bar{I}_b$ is the average grey-scale value in neighboring area around the target.

In this experiment, the moving target is a circle with diameter $d=3^{\circ}$, and the velocity is $v=300^{\circ}/s$. Different contrast is chosen by adjusting the target color. We compare the mean neural response under different kernel sizes $M_W$ and standard deviations $\sigma_W$, since WACs regulate the balance of excitation/inhibition directly, which also can affect the degree of sensitivity. Other parameters are the same as the ones in the former experiment. The response intensity of DSGCs with respect to the contrast is plotted in Fig. \ref{fig:s04f04a} and Fig. \ref{fig:s04f04b}. As the contrast increases, the outputs of our model grow approximately quadratically. When $c=1$, the outputs reach maximum. Meanwhile, the antagonistic center-surround filter is also able to affect output intensity directly. The larger kernel size and standard deviation generate a larger excitation/inhibition ratio, which causes the increase of response intensity.

Furthermore, we compare the contrast sensitivity of BRNN with other typical bioinspired models, including ESTMD \cite{wiederman2008model}, IIT \cite{bagheri2017performance}, DSTMD \cite{wang2020a}. The normalized responses of them are plotted in Fig. \ref{fig:s04f05a}. It can be seen that they have similar contrast sensitivity. As the contrast increases, the response intensity increases approximately quadratically.

\subsubsection{Size Sensitivity}
Another important attribute is size sensitivity. Biological experimental results in \cite{Hoggarth2015} demonstrate that the amplitudes of spiking responses in RGCs increase more sharply with the target diameter up to an optimal size, and decline sharply when the target size beyond an optimal size. It results from the wide-field suppressive surround mechanism. In our proposed model, kernel sizes $M_W$ and standard deviations $\sigma_W$ are two main important variables to control the wide-field suppression. Therefore, we investigate the size sensitivity with different kernel sizes and standard deviations. The target size is set as $d=3^{\circ}$ and the velocity is $v=300^{\circ}/s$ in default.

Fig. \ref{fig:s04f04c} shows the size tuning with different kernel sizes, where standard deviation $\sigma_W$ is fixed as $1.2$. Obviously, mean neural response amplitude reaches the peak at an optimal size ($2^{\circ}\sim 4^{\circ}$), which is consistent with the biological data \cite{Hoggarth2015}. Meanwhile, the increased kernel size can enhance response amplitude and enlarge the optimal size. One explanation is that the optimal size closely matched the dendrite field size \cite{Hoggarth2015}, which is positively related to kernel size. Fig. \ref{fig:s04f04d} plots the curve of size sensitivity with different $\sigma_W$ and a fixed kernel size $M_W=15$. Likewise, the increased standard deviation is also able to enhance response amplitude and enlarge the optimal size gradually but with a decreasing rate. Under a limited dendrite field size (fixed kernel size), the standard deviation actually can finely adjust the size sensitivity by changing the excitation/inhibition ratio.

The size tuning of BRNN, ESTMD, IIT and DSTMD is shown in Fig. \ref{fig:s04f05b}. In this experiment, the optimal sizes of all models concentrate between $1^{\circ}\sim 3^{\circ}$, which demonstrates that they are more sensitive to small targets. Meanwhile, by adjusting $M_W=11$ and $\sigma_W=1.2$, our proposed model shows the size tuning similar to the ESTMD. As an advantage, BRNN has a larger range of size sensitivity by choosing different $M_W$ and $\sigma_W$.

\subsubsection{Velocity Sensitivity}
Velocity sensitivity is another basic property of motion vision \cite{wyatt1975directionally}. The neural response is usually strongest at a specific velocity of movement and tuned across a broad range of velocities. If the target moves too slowly, motion information will be difficult to be perceived. On the contrary, if the target moves too fast, many motion information will be lost due to relatively slow neural computing. In like manner, we also investigate the velocity sensitivity with different kernel sizes and standard deviations. The target diameter is set as $d=3^{\circ}$ and the Weber contrast is $c=1$.

\begin{figure*}[!htbp]
    \centering
    \subfigure[]{\label{fig:s04f04a}\includegraphics[width=0.3\textwidth]{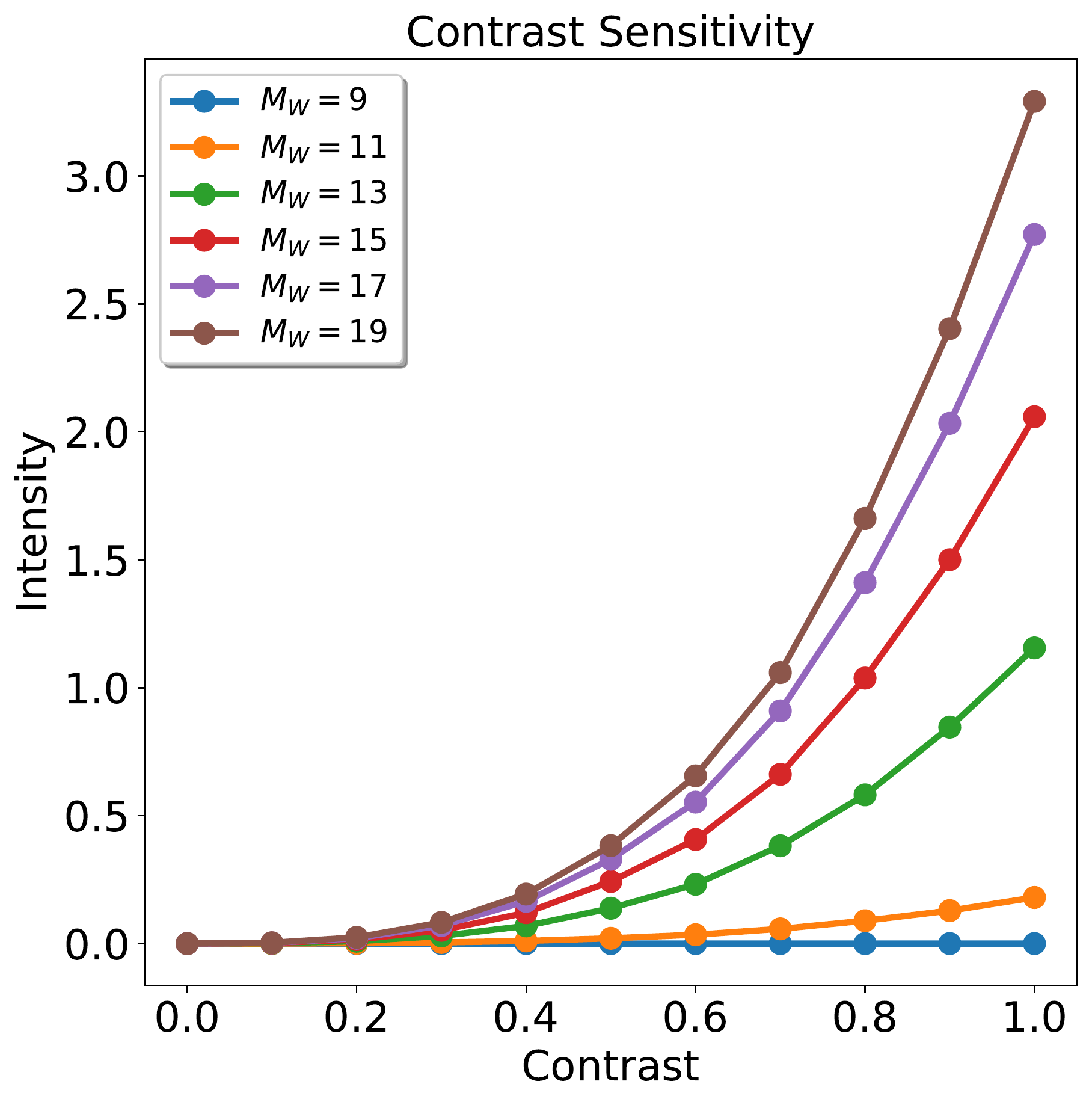}}%
    ~\hspace{2pt}
    \subfigure[]{\label{fig:s04f04b}\includegraphics[width=0.3\textwidth]{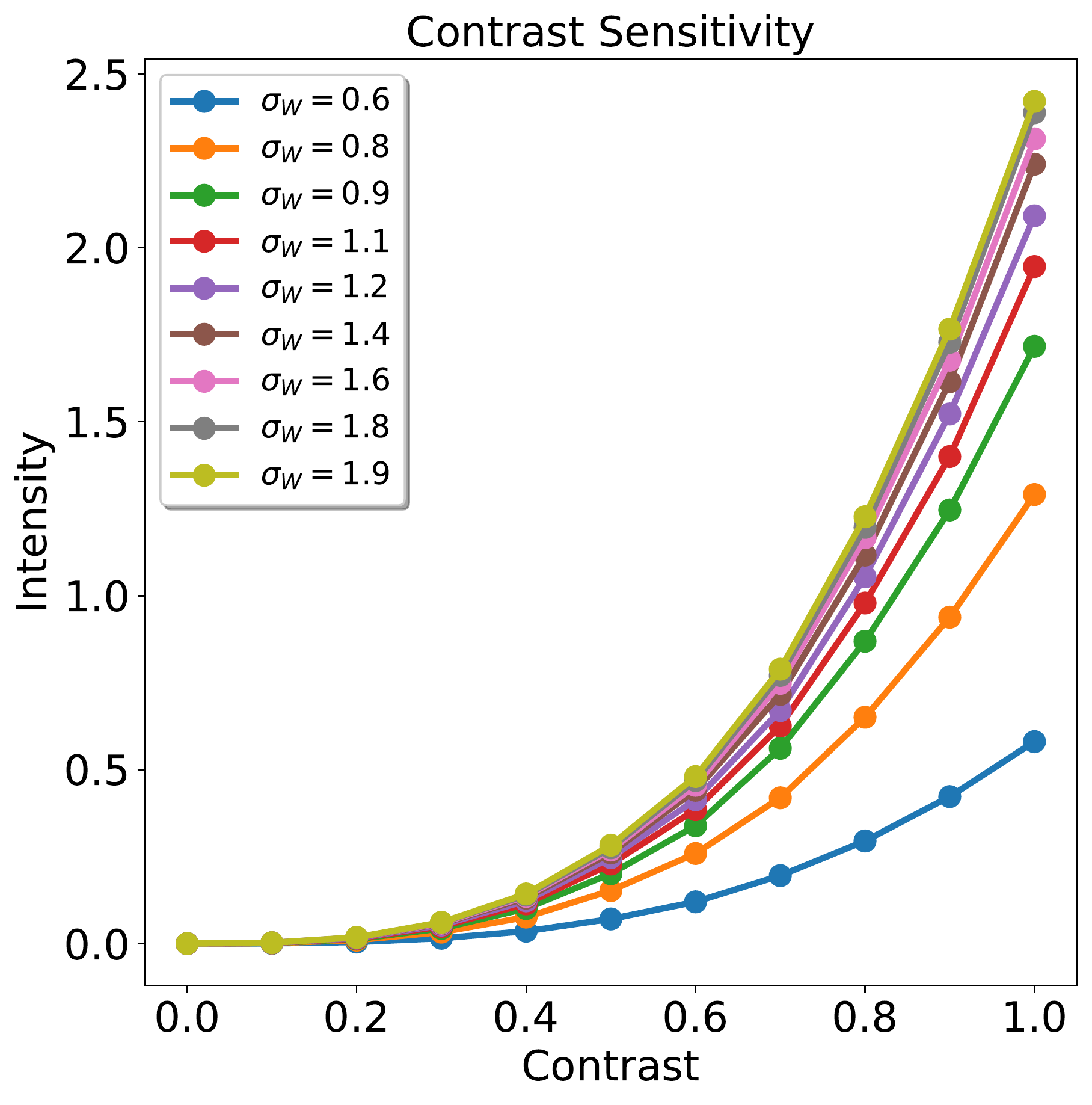}}
    ~\hspace{2pt}
    \subfigure[]{\label{fig:s04f04c}\includegraphics[width=0.3\textwidth]{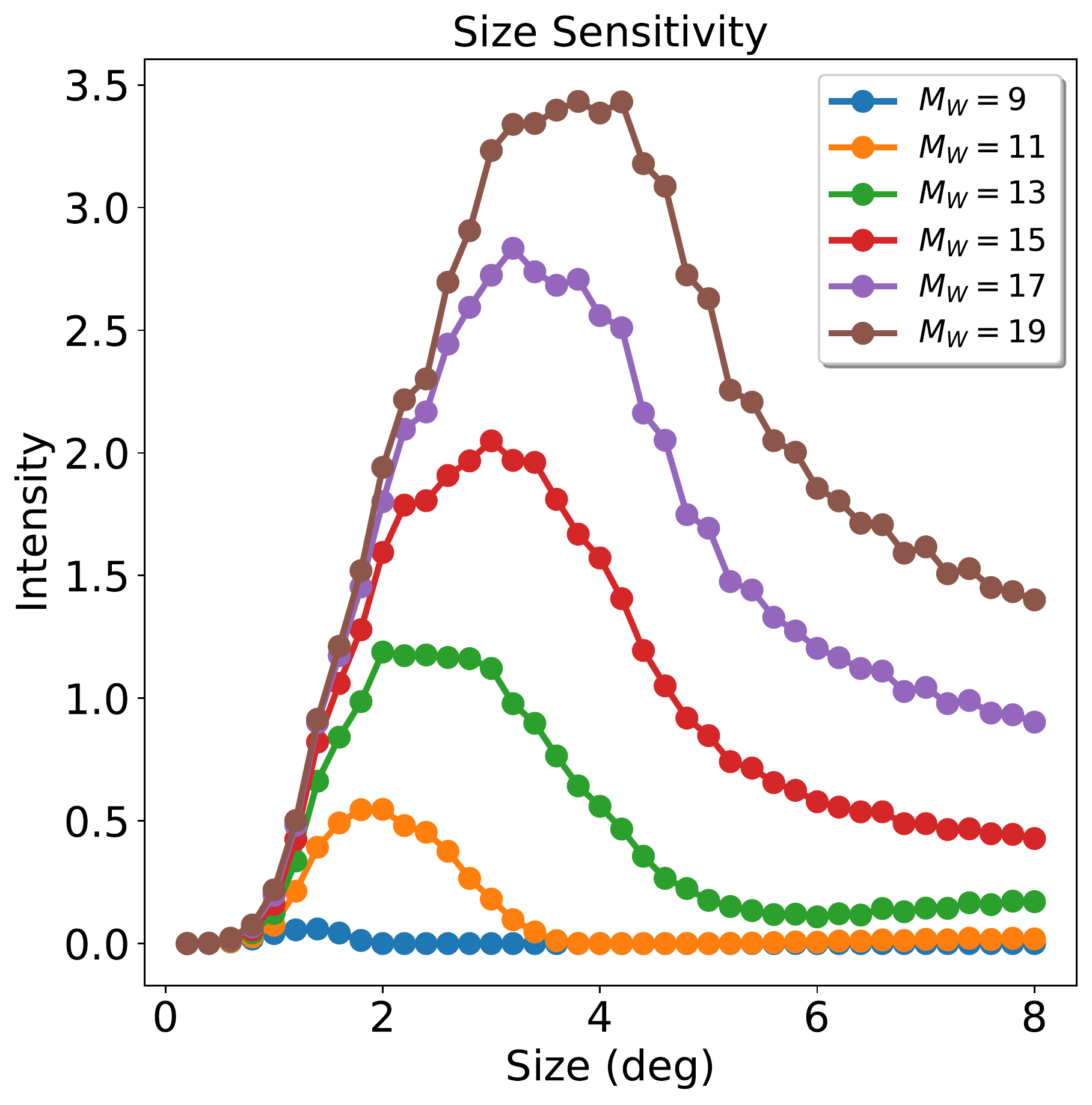}}\\
    \subfigure[]{\label{fig:s04f04d}\includegraphics[width=0.31\textwidth]{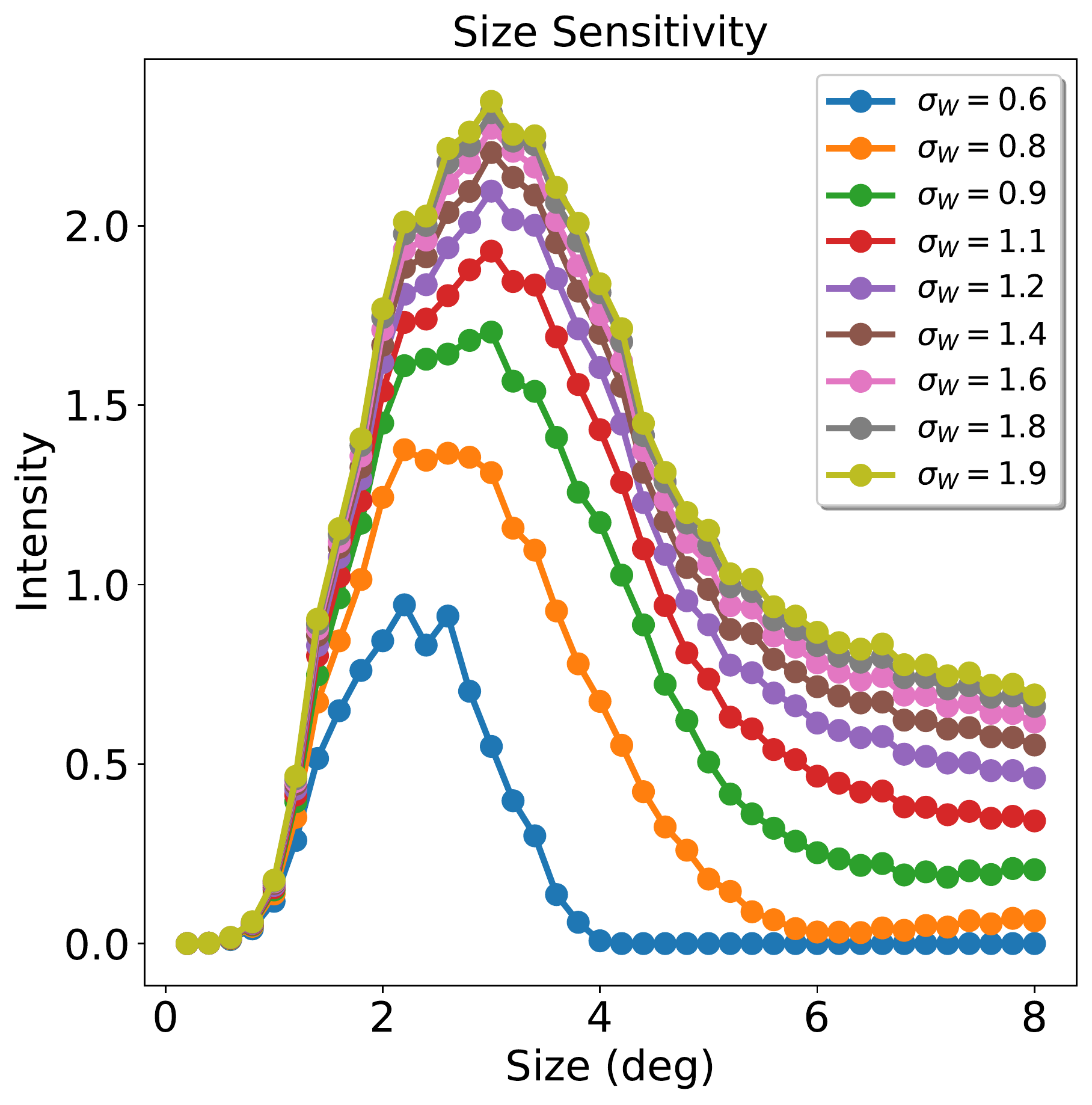}}%
    ~\hspace{2pt}
    \subfigure[]{\label{fig:s04f04e}\includegraphics[width=0.3\textwidth]{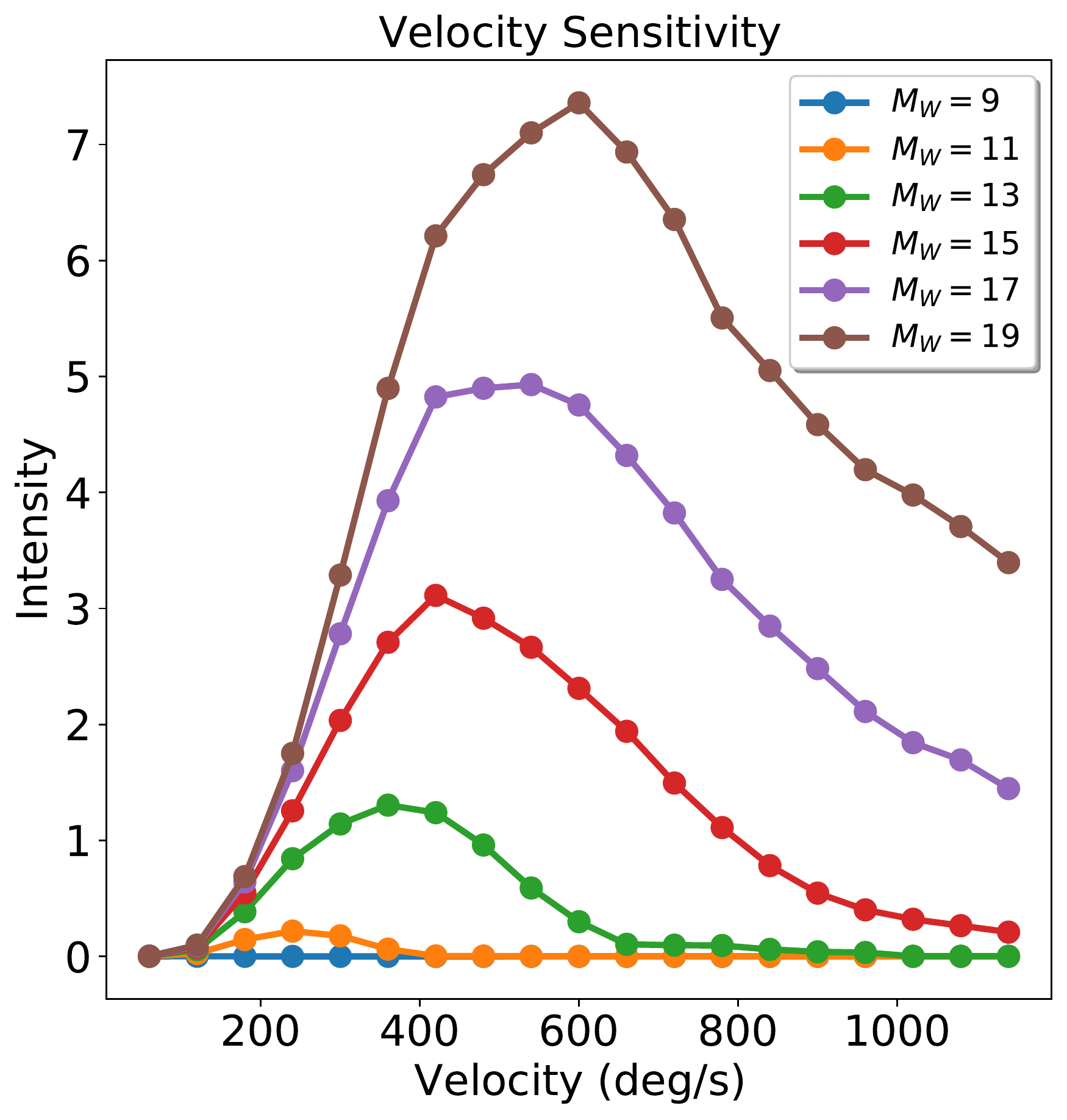}}
    ~\hspace{2pt}
    \subfigure[]{\label{fig:s04f04f}\includegraphics[width=0.31\textwidth]{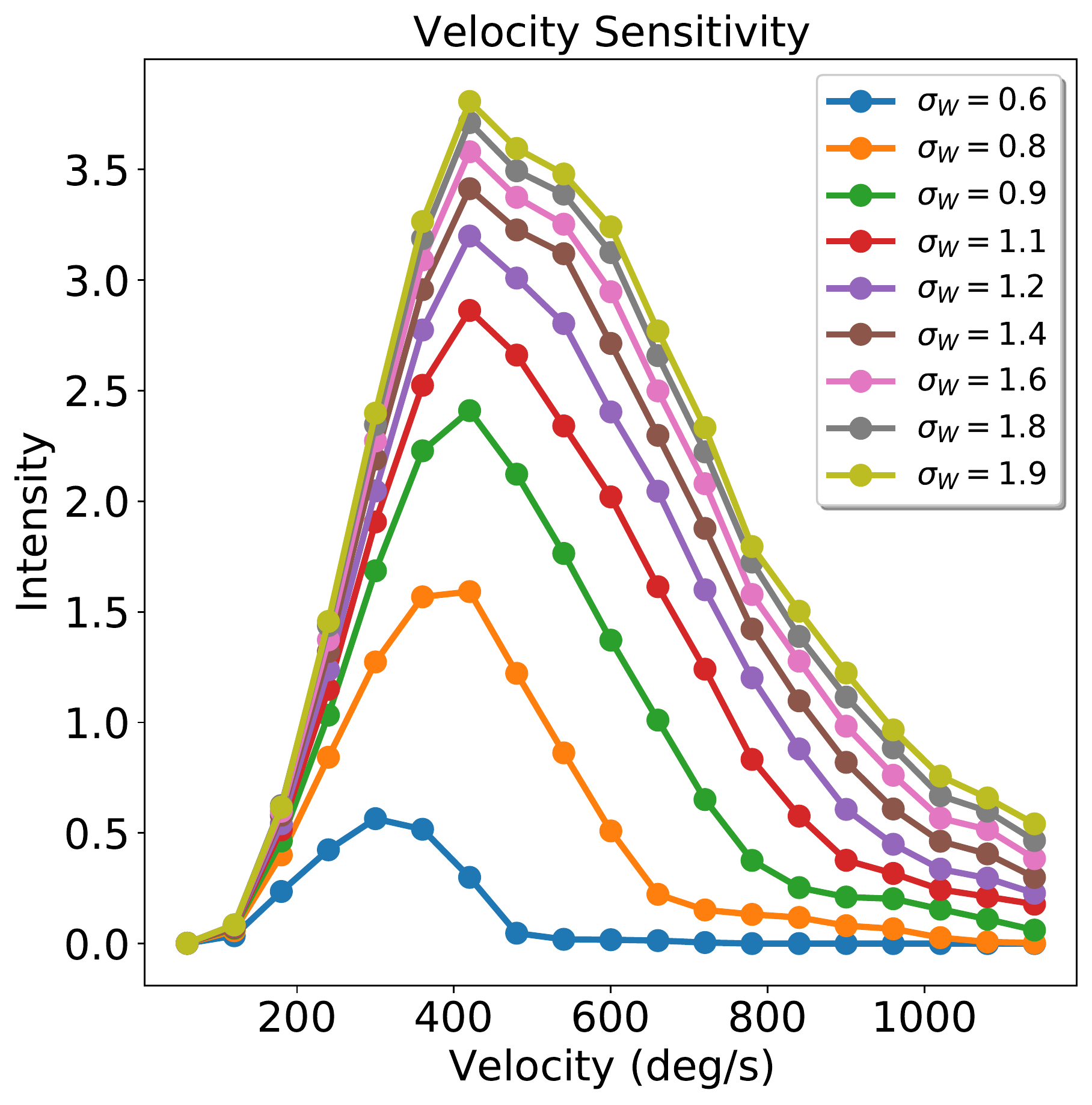}}
    
    \caption{The contrast, size and velocity sensitivity of the BRNN under different conditions. (a) The contrast tuning at different kernel size $M_W$ ($\sigma_W=1.2$). (b) The contrast tuning at different $\sigma_W$ ($M_W=15$). (c) The size tuning at different kernel size $M_W$ ($\sigma_W=1.2$). (d) The size tuning at different $\sigma_W$ ($M_W=15$). (e) The velocity tuning at different kernel size $M_W$ ($\sigma_W=1.2$). (f) The velocity tuning at different $\sigma_W$ ($M_W=15$).}
    \label{fig:s04f04}
\end{figure*}

The neural responses of DSGCs are plotted in Fig. \ref{fig:s04f04e} with changing kernel sizes. Obviously, the increasing velocity can strengthen the response amplitude before an optimal velocity ($200\sim 600^{\circ}/s$). When the target moves beyond the optimal velocity, the response intensity declines gradually. The increased kernel size can also enhance response amplitude and enlarge the optimal velocity, similar to the size sensitivity. We further fix $M_W=15$ and investigate the effect of $\sigma_W$ on the output. The result is shown in Fig. \ref{fig:s04f04f}, which demonstrates that the standard deviation is able to finely tune the optimal velocity with the constraint of dendrite field size.

\begin{figure*}[!htbp]
    \centering
    \subfigure[]{\label{fig:s04f05a}\includegraphics[width=0.3\textwidth]{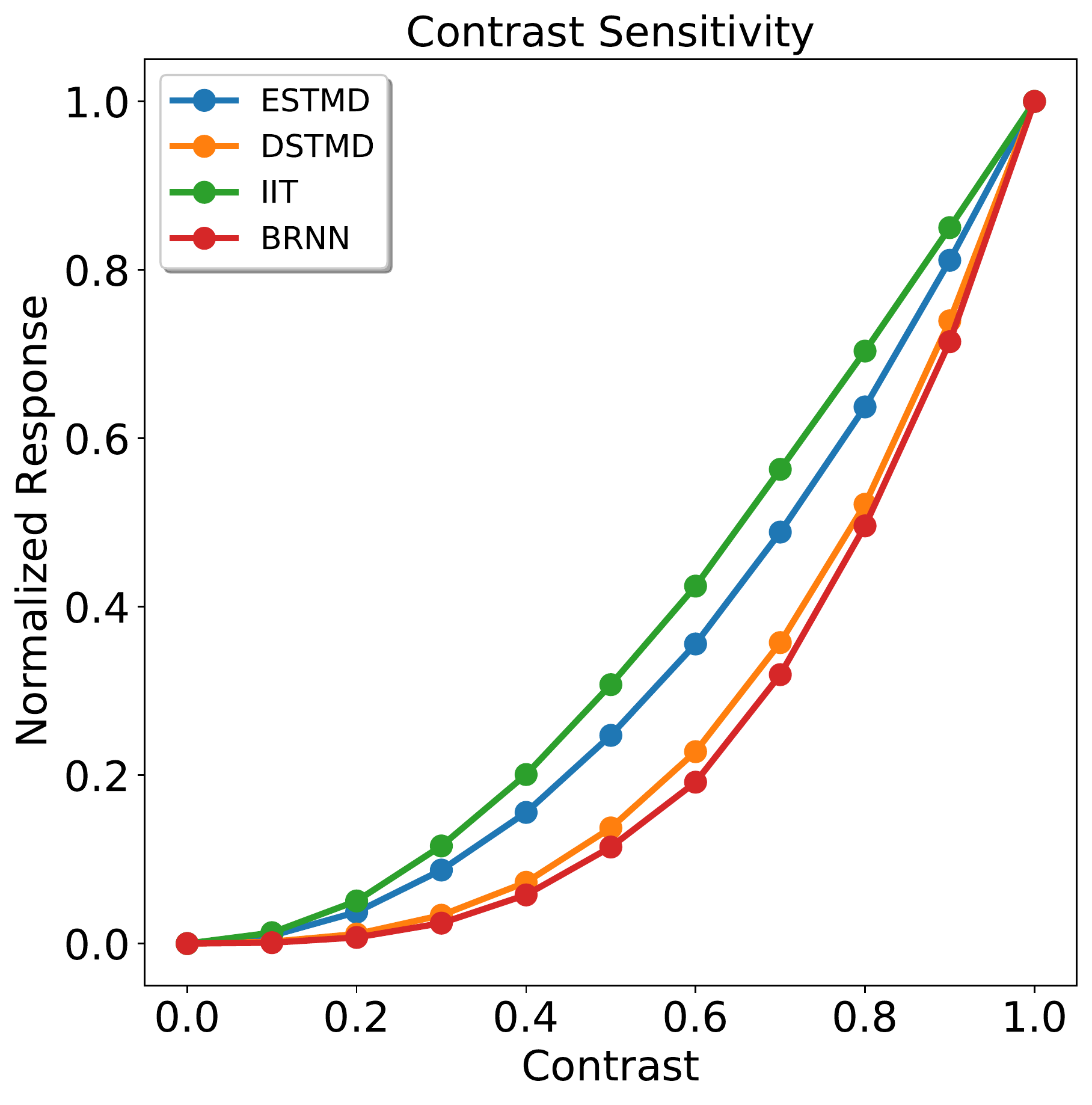}}%
    ~\hspace{2pt}
    \subfigure[]{\label{fig:s04f05b}\includegraphics[width=0.3\textwidth]{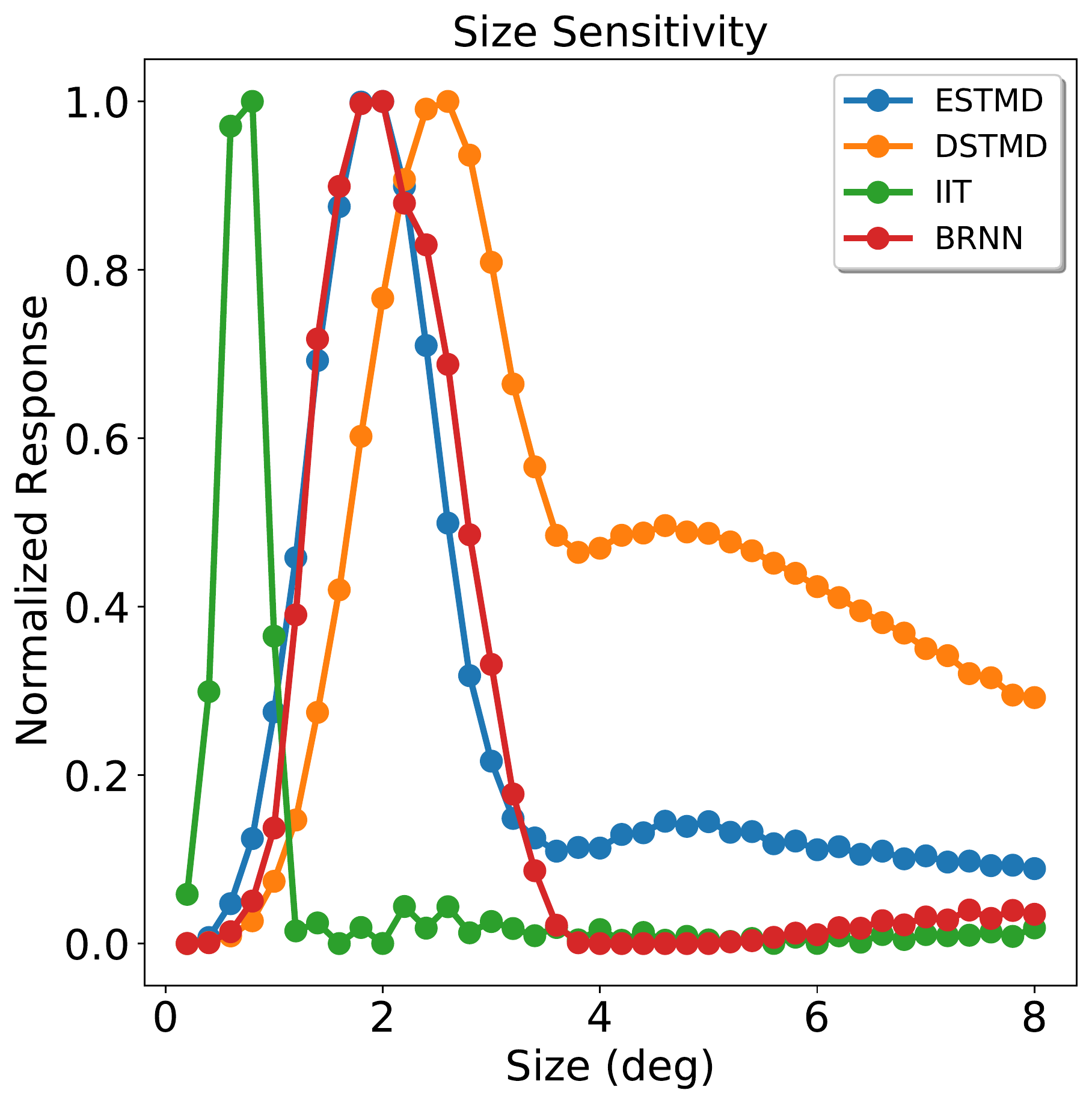}}
    ~\hspace{2pt}
    \subfigure[]{\label{fig:s04f05c}\includegraphics[width=0.3\textwidth]{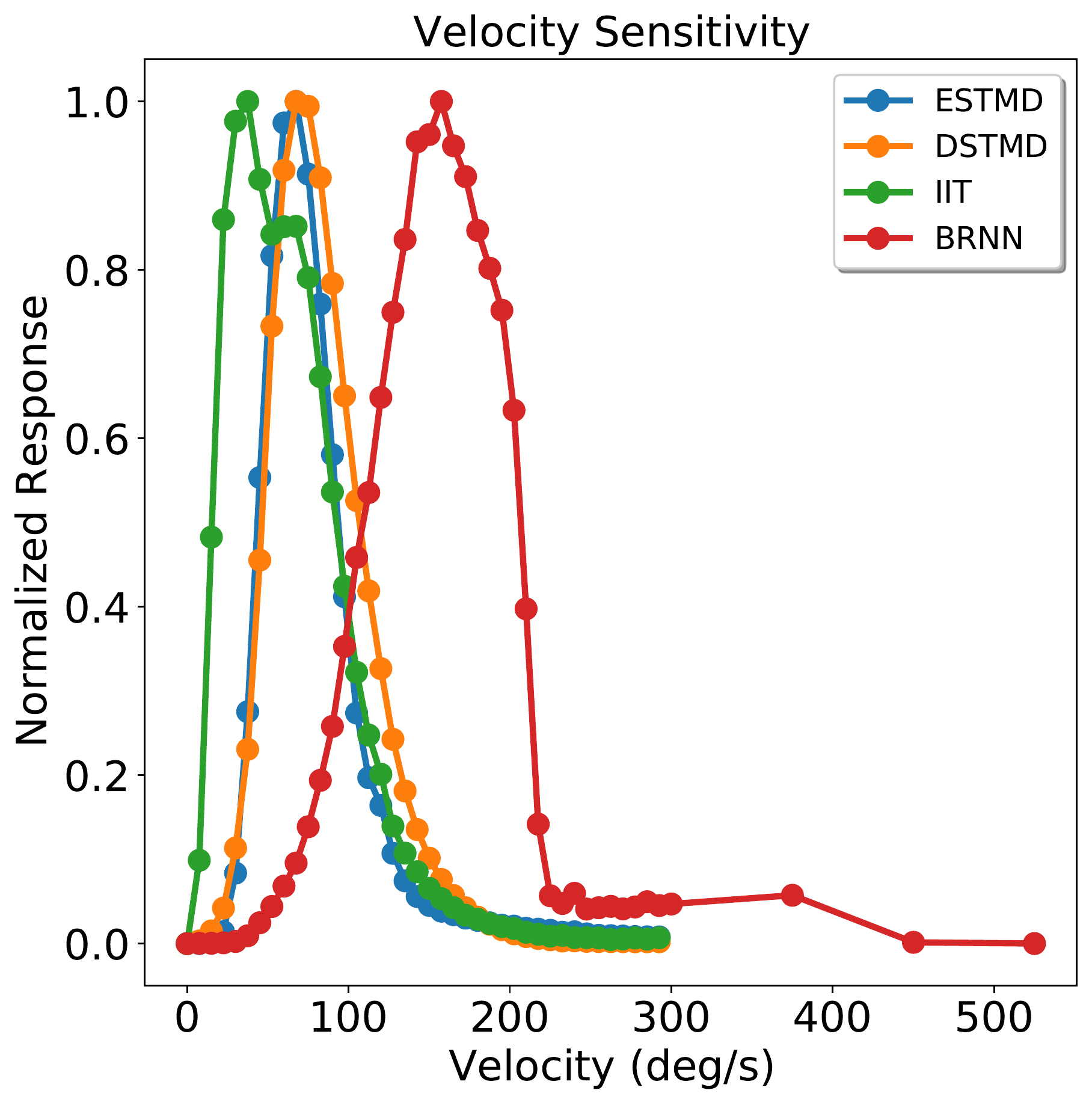}}\\
    \caption{The contrast, size and velocity sensitivity of the BRNN compared with other typical bioinspired models. (a) The normalized mean responses with different contrast. (b) The normalized mean responses with different sizes. (b) The normalized mean responses with different velocities.}
    \label{fig:s04f05}
\end{figure*}

We find that the optimal velocities of other models are generally lower ($50^{\circ} \sim 100^{\circ}/s$), as shown in Fig. \ref{fig:s04f05c}. DSTMD also proposed that the optimal velocity could be tuned by adjusting the parameters of temporal filter. While in comparison, the proposed model has a larger range of velocity sensitivity by changing $M_W$ and $\sigma_W$. In Fig. \ref{fig:s04f05c}, the optimal velocity of BRNN is $\sim 150^{\circ}/s$ when $M_W=11,\sigma_W=0.7$.

\subsubsection{Response Speed}
\begin{figure}[!htbp]
    \centering
    \includegraphics[width=0.4\textwidth]{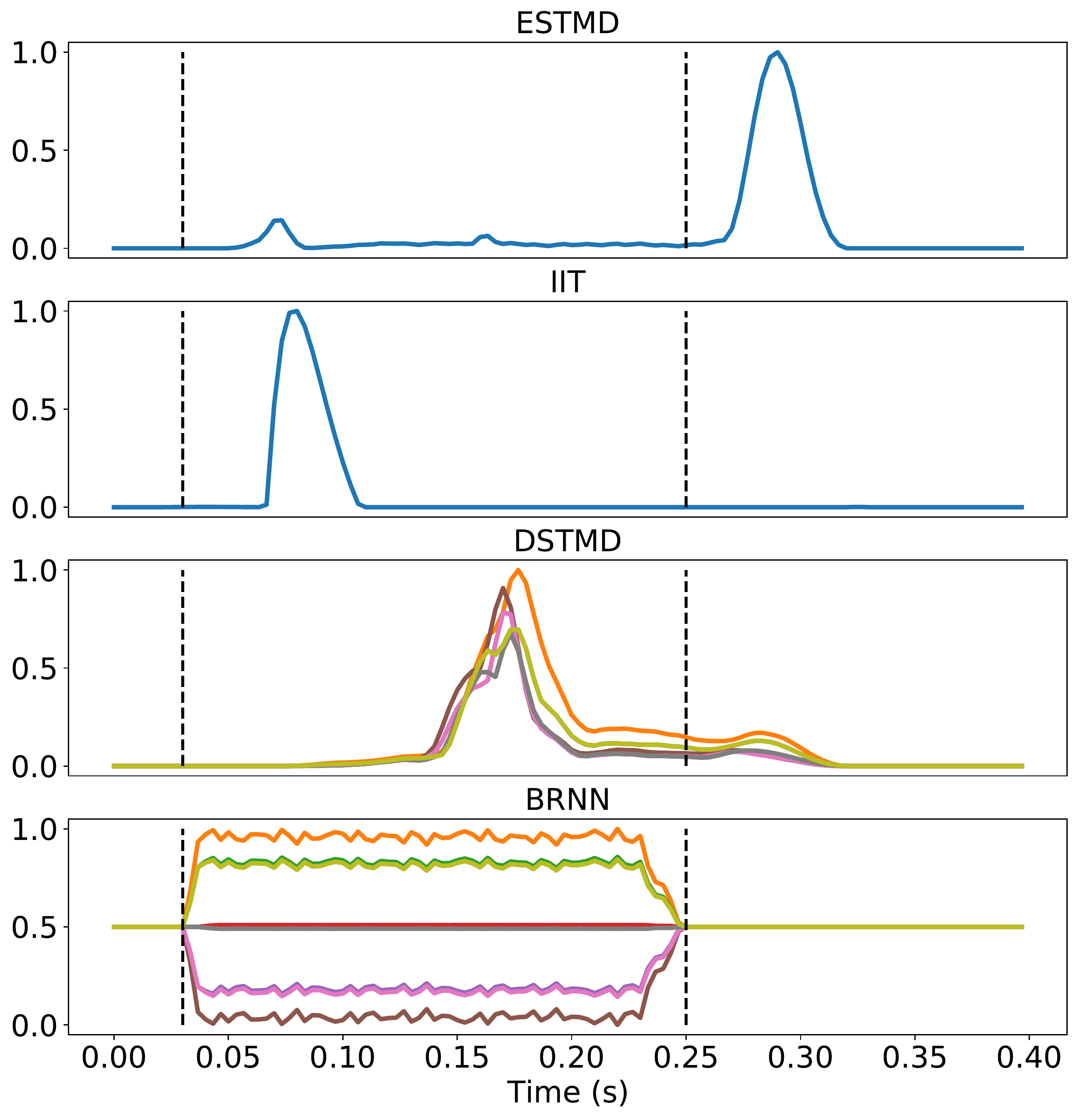}
    \caption{The normalized outputs of four models at  $d=2^{\circ}, \; v=75^{\circ}/s$.}
    \label{fig:s04f06}
\end{figure}

The typical temporal band-pass filter usually suffers from phase delay or phase shift in the discrete digital filtering process, which has a strong impact on accuracy and realtime performance. It needs to store a period of history and implement filtering in a moving manner. In this paper, the proposed neurodynamics-based temporal filter can achieve the same function as the traditional band-pass filter in essence. An advantage is that the proposed filter is easy to reduce phase delay through selecting the superficial neural outputs, while without affecting the filtering performance. It actually encodes the history of original signal into the neuronal activities in different layers, which converts temporal computation into spatial computation. The normalized outputs of four models are plotted in Fig. \ref{fig:s04f06}, where a small target moves at a constant speed from $0.03-0.25s$. The ESTMD, IIT and DSTMD have obvious phase delay such that the response speed is relatively slow, the response time is about $40ms$. While the proposed temporal filtering enables the model to respond to motion rapidly.

\subsection{Motion Detection in A Synthetic Scenario}
In this section, the ability of motion detection in cluttered backgrounds is tested. The process of motion detection is described in Section \Rmnum{3}-D. A receiver operating characteristic (ROC) curve is plotted to illustrate the diagnostic ability of the proposed motion detector. This curve is created by plotting the true-positive rate (TPR) against the false-positive rate (FPR) at various threshold $\gamma$ settings. The true-positive rate represents the probability of successful detection, which is calculated by the number of true detections $N_{TD}$ and the number of actual targets $N_{AT}$.
\begin{equation}
TPR = \frac{N_{TD}}{N_{AT}}.
\end{equation}
The false-positive rate represents the probability of false alarm. The total number of time steps is fixed at $N_T$, we set the number of false positives $FPR$ as abscissa, defined as follows.
\begin{equation}
FPR = \frac{N_{FP}}{N_T}.
\end{equation}

\begin{figure}[!htbp]
    \centering
    \includegraphics[width=0.5\textwidth]{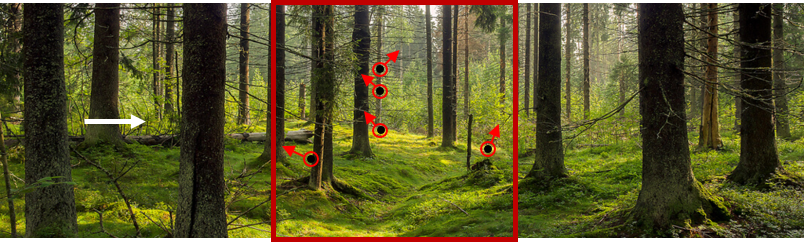}
    \caption{The scene of multiple small targets moving in a cluttered background.}
    \label{fig:s04f08}
\end{figure}

The synthetic environment is shown Fig. \ref{fig:s04f08}, where the background is cluttered and dark. The cluttered backgrounds moves from left to right at different speed. Five targets move from random initial positions in five random directions respectively. If two targets collide or any target is beyond the boundary, they will move in the opposite direction. Six groups of experiments are conducted to investigate the influences of luminance, size, velocity of targets, velocity of background and background suppression algorithm. By default, the total number of time steps is $180$ and the frame rate is $300Hz$. The luminance, diameter and velocity of targets are set to $l=0$, $d=2^{\circ}$ and $v=300^{\circ}/s$ respectively. The velocity of background is in default set to $v_b=75^{\circ}/s$. We use the DBSCAN algorithm in scikit-learn to discover the clusters of the regions of interest. There are two most important parameters in this function, $\epsilon$ and $N_{min}$. $\epsilon$ represents the maximum distance between two samples for one to be considered as in the neighborhood of the other. $N_{min}$ is the number of samples in a neighborhood for a point to be considered as a core point. Herein, $\epsilon$ and $N_{min}$ are set to 3 and 1 respectively. If the distance between the clustering center and the actual target position less than a threshold $d_{th}=0.5d+1^{\circ}$, the target is considered to be detected successfully.

\begin{figure*}[!htbp]
    \centering
    \subfigure[]
    {\label{fig:s04f09a}\includegraphics[width=0.3\textwidth]{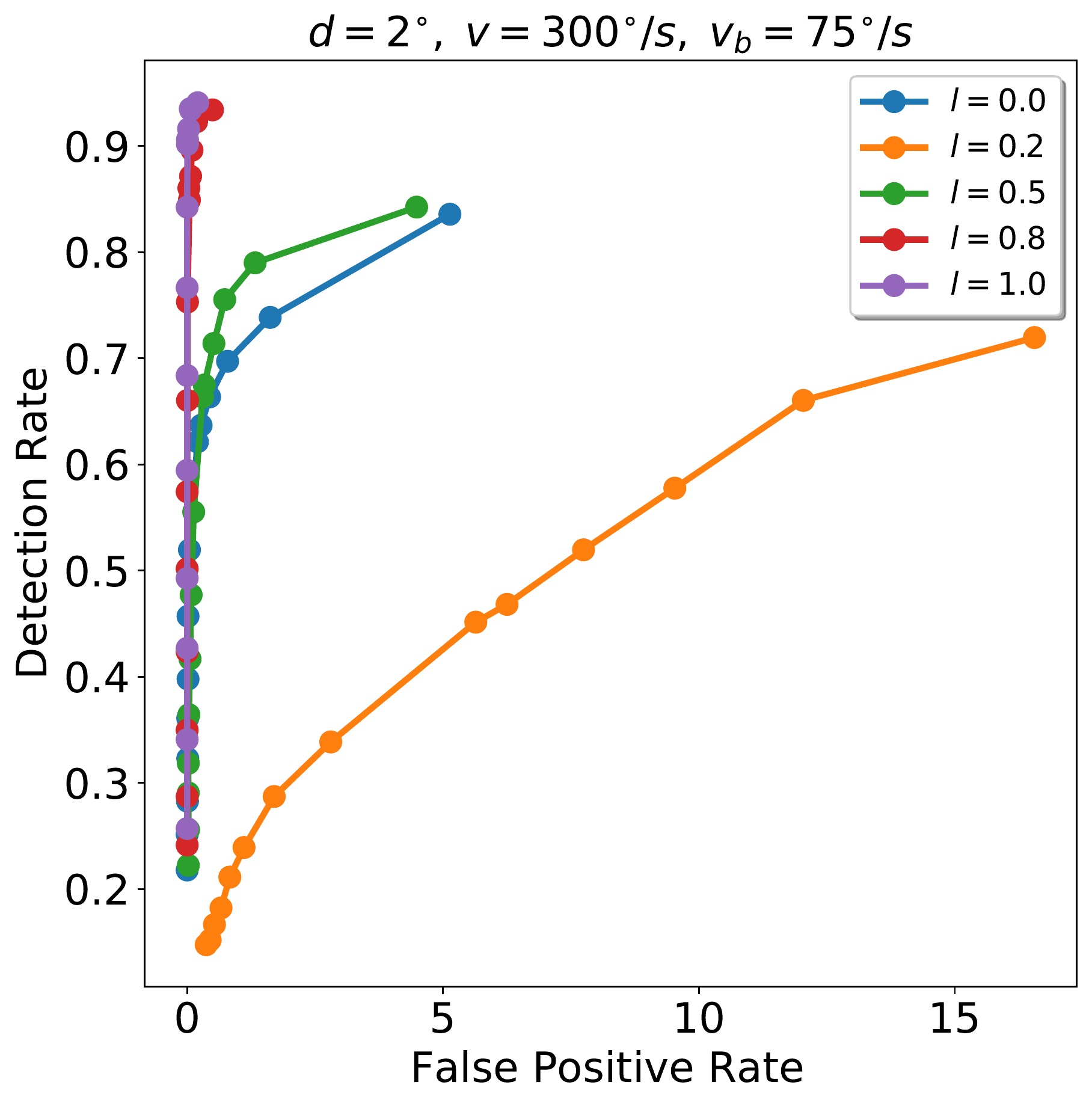}}%
    ~\hspace{2pt}
    \subfigure[]{\label{fig:s04f09b}\includegraphics[width=0.3\textwidth]{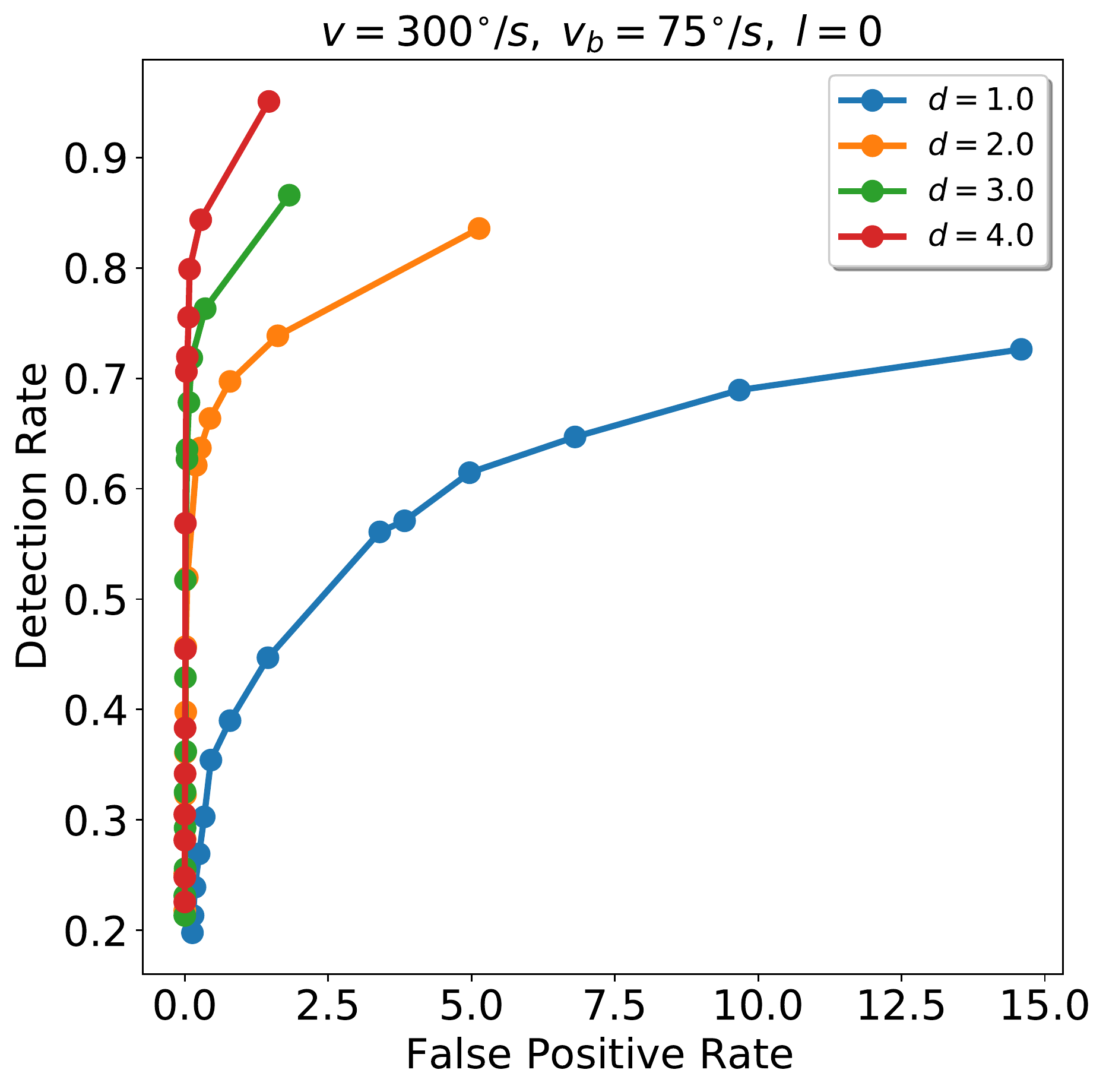}}
    ~\hspace{2pt}
    \subfigure[]{\label{fig:s04f09c}\includegraphics[width=0.3\textwidth]{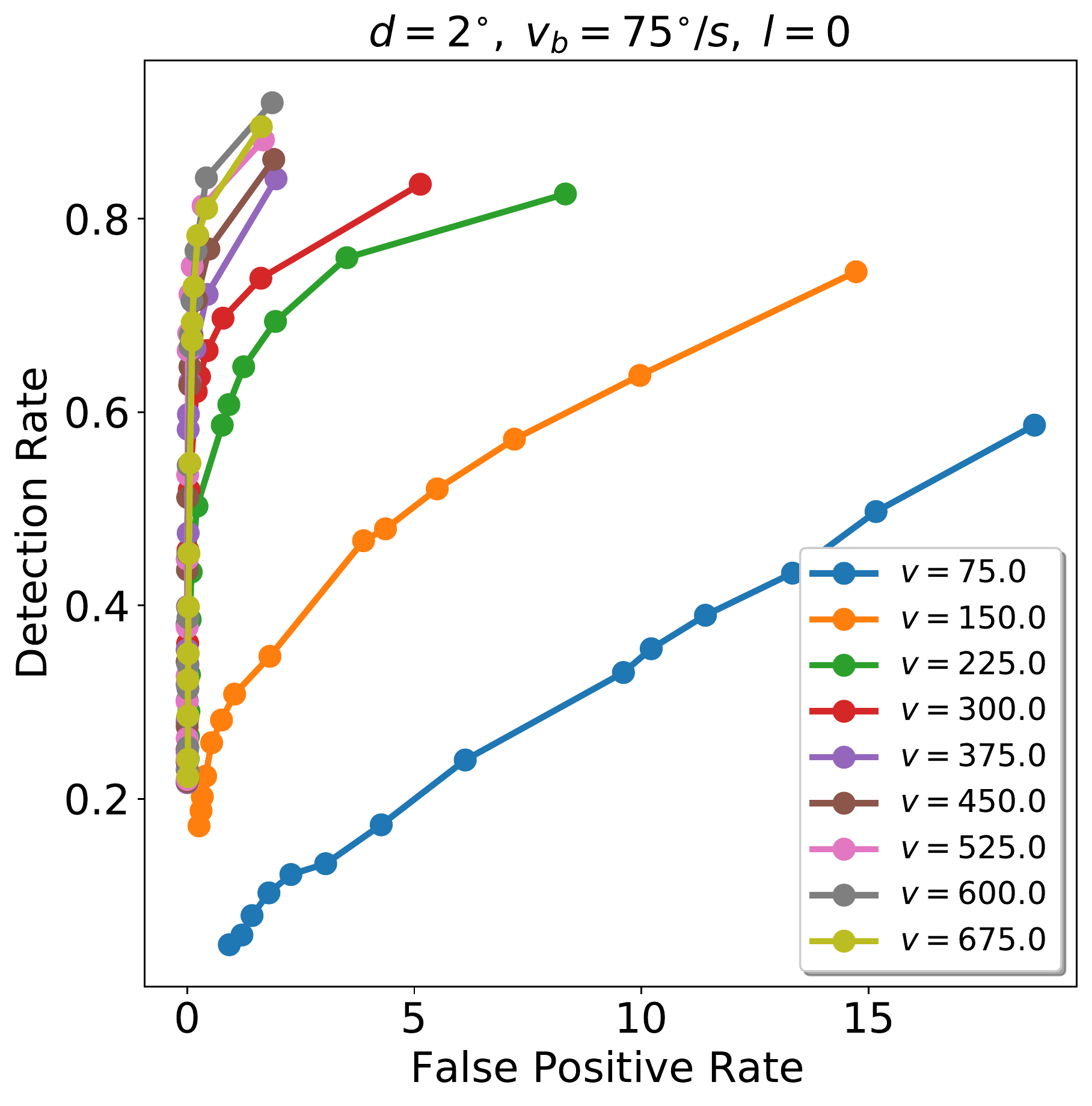}}\\
    \subfigure[]
    {\label{fig:s04f09d}\includegraphics[width=0.3\textwidth]{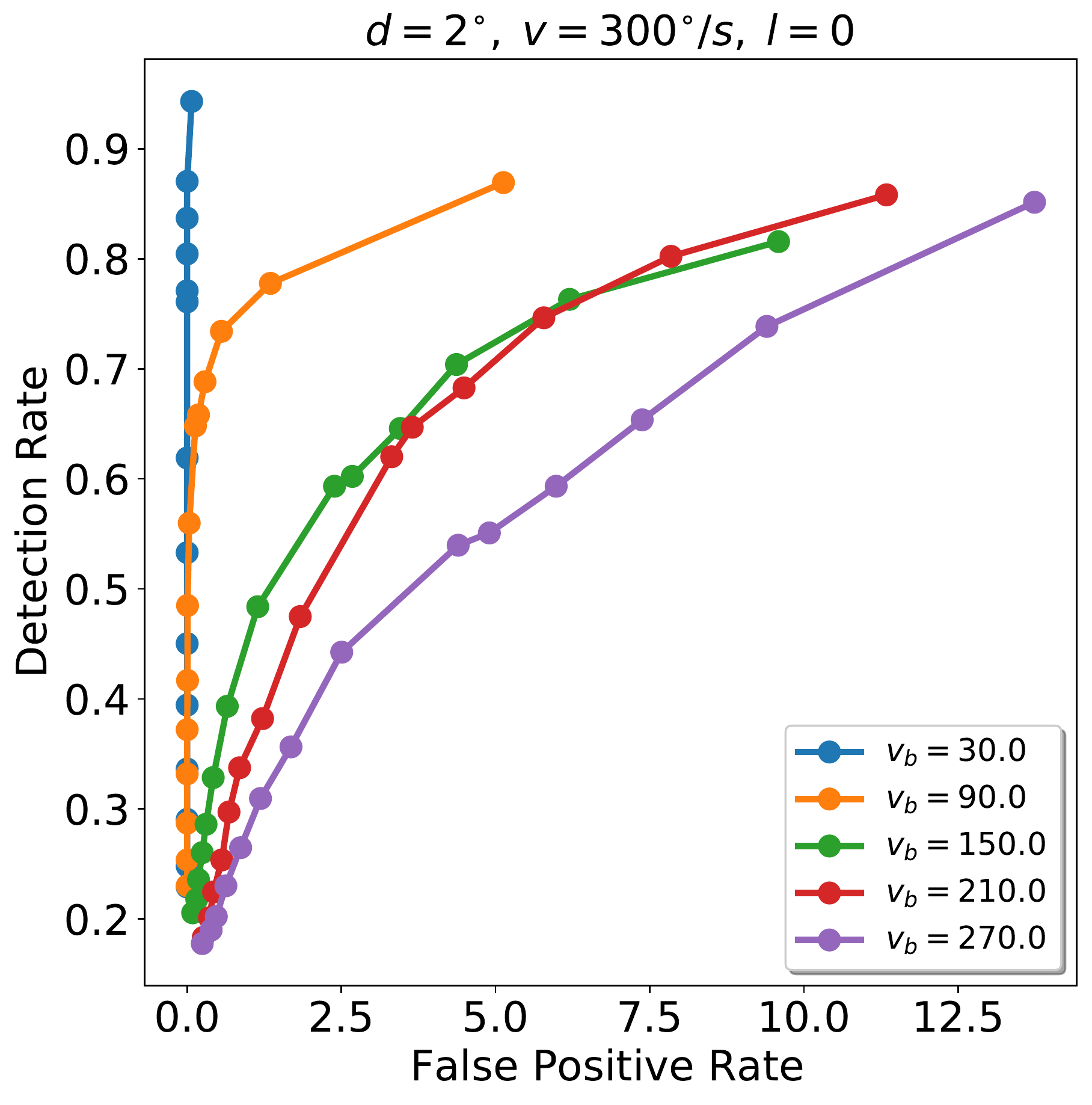}}%
    ~\hspace{2pt}
    \subfigure[]{\label{fig:s04f09e}\includegraphics[width=0.3\textwidth]{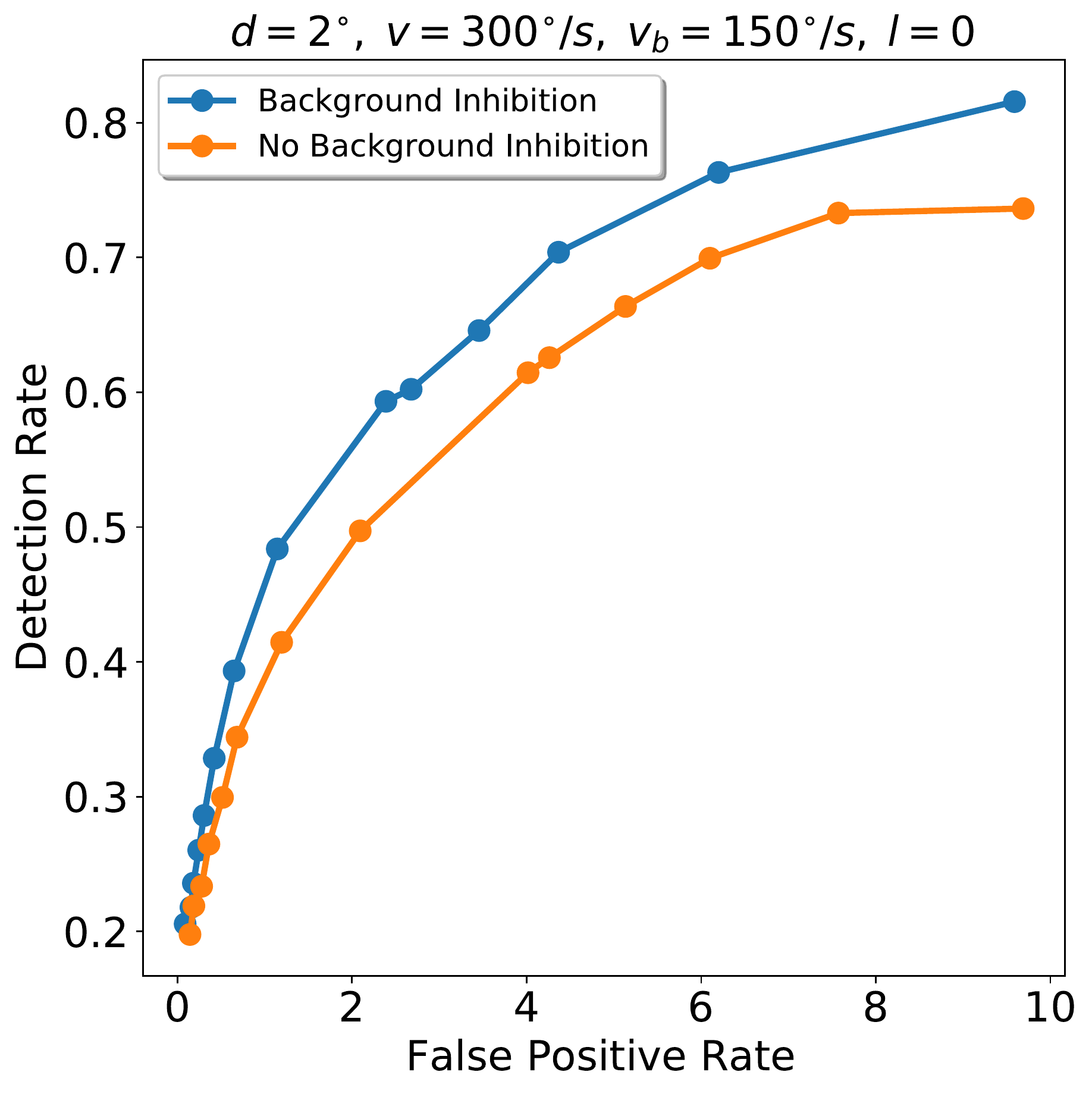}}
    ~\hspace{2pt}
    \subfigure[]{\label{fig:s04f09f}\includegraphics[width=0.3\textwidth]{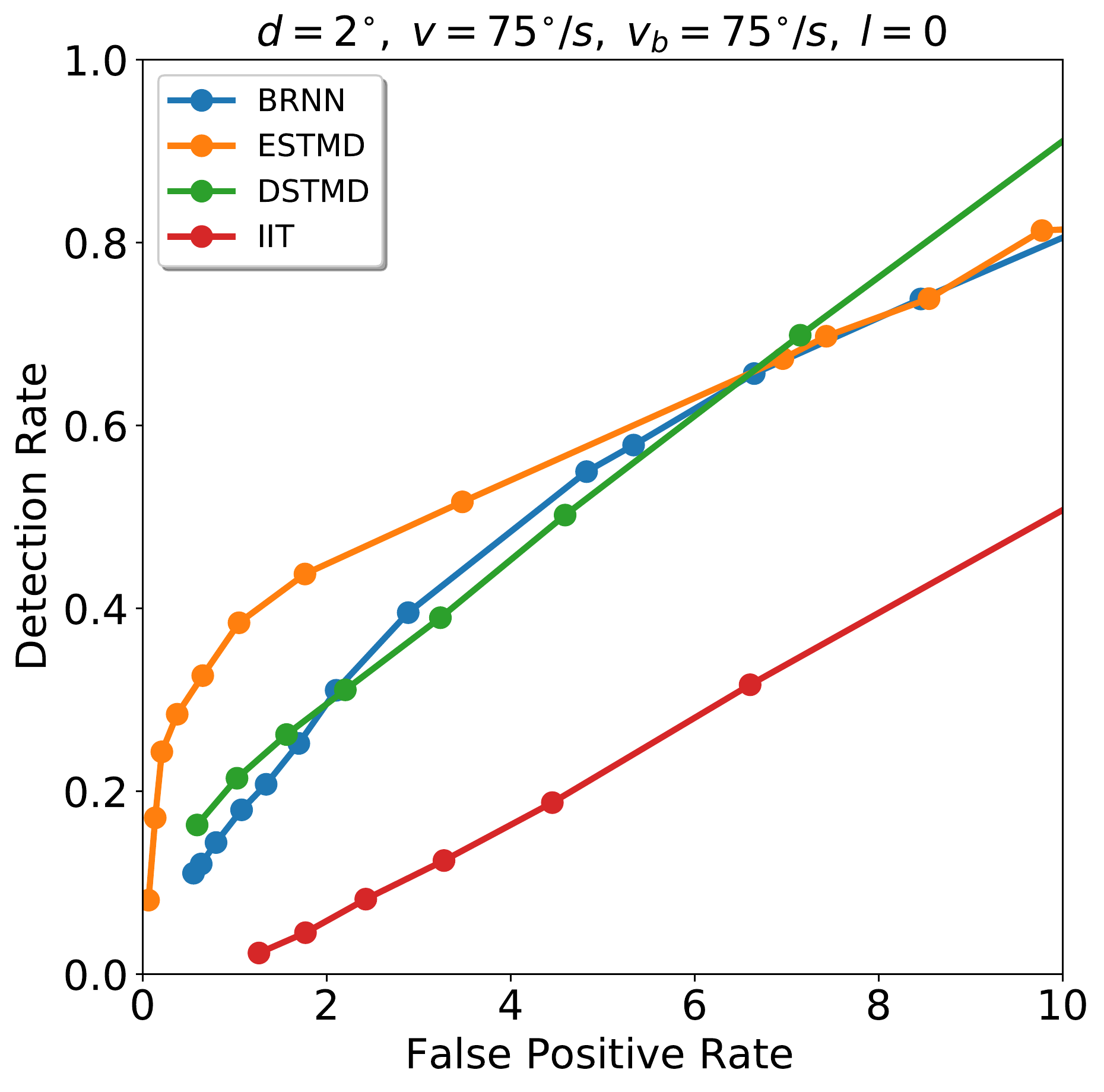}}
    \caption{The ROC curves of the BRNN for motion detection under different settings.}
    \label{fig:s04f09}
\end{figure*}

We firstly choose $M_W=21$ and $\sigma_W=1.5$, and evaluate the diagnostic ability of the proposed motion detector with different target luminance. In this group, the luminance of targets is set to $I=0.0, 0.2, 0.5, 0.8, 1.0$ respectively, where the actual RGB value is $[255I,\; 255I,\; 255I]$. The ROC curve is plotted in Fig. \ref{fig:s04f09a}. There we can see that the detection performance is lowest when the luminance is $0.2$, and good performance occurs when targets are bright. This is because the luminance of background is close to $0.2$ such that the Weber contrast is low when $l=0.2$. While the BRNN shows a good performance of motion detection when targets are very dark or bright, since the Weber contrast is relatively high in these situations. It also illustrates that the proposed model is sensitive to the visual change in both ON and OFF channels.

The Fig. \ref{fig:s04f09b} shows the detection performance of the BRNN with different target sizes. The sizes of targets are set to $d=1^{\circ}$, $d=2^{\circ}$, $d=3^{\circ}$ and $d=4^{\circ}$ respectively. As the size increases, the detection rate of the BRNN also increases for a given false positives. A relatively large target can be detected more easily. The proposed model is sensitive to the targets of different sizes around the optimal size. Meanwhile, the optimal size can be adjusted by changing the parameters of antagonistic center-surround spatial filtering, which can expand the range of size sensitivity. By contrast, the previous typical models are only sensitive to a small range of size (Fig. \ref{fig:s04f05b}), which actually limits their detection ranges.

The Fig. \ref{fig:s04f09c} shows the ROC curve of the BRNN with different motion velocities. The velocities of targets are set to $75^{\circ}/s \sim 675^{\circ}/s$. Obviously, the detection ability of the BRNN is enhanced as the motion velocity increases, when it is lower than the optimal velocity. If $v>600^{\circ}/s$, the ability of motion perception will diminish . The proposed BRNN is effective to detect the targets of different velocities around the optimal velocities. Meanwhile, the optimal value can be changed by changing the parameters of antagonistic center-surround spatial filtering, which can also expand the speed sensitive range of small targets.

The influence of background velocity is further investigated in the fourth experiment. As shown in Fig. \ref{fig:s04f09d}, fast background movement can weaken the capability of small-target motion detection. It is not difficult to understand that a faster moving background produces many small target-like features which will increase false positive rate greatly. In addition, the directionally selective inhibitioin algorithm can effectively reduce the interference from the moving background in our model, which is reflected in Fig. \ref{fig:s04f09e}. In the situation with backgound suppression, the threshold $N_{th}$ is set to $6$. For a given false positive rate, the process of direction selective inhibitioin can improve the detection rate greatly. This is significant for robotic visual system to overcome the target detecting problem with an instable visual input.

Finally, we also compare the performance of our BRNN with the ESTMD, DSTMD and IIT at the optimal value of these classical models, where $d=2^{\circ}, v=75^{\circ}/s$. In order to adapt to small and slow moving targets, we set the kernel size and standard deviation to $M_W=11,\sigma_W=0.7$. The ROC curve is plotted in Fig. \ref{fig:s04f09f}. The ESTMD and DSTMD are relatively good at detecting slow motion. The proposed BRNN is also able to obtain similar detection performance. But it is worth noting that sensing slow movement usually needs high sampling frequency, which may be difficult to perform in low-cost hardware systems.

\subsection{Motion Detection in Real Scenarios}
The proposed model is further tested in the STNS dataset \cite{bagheri2017performance}, which contains small moving targets within several real cluttered natural environments. All frames are used to evaluate the diagnostic ability of the proposed motion detector. All images are resized to one-half as $320\times 240$. The FOV for this dataset is also not available, therefore we assume FOV $\sim80^{\circ}$. About the DBSCAN algorithm,  $N_{min}$ and $\epsilon$ are set to 8 and 2 respectively. Additionally, the distance threshold $d_{th}$ is set to $0.5d+1^{\circ}$, where $d$ represents the diagonal length of the target. For each scenario, the detection threshold $\gamma$ is chosen as $0.01-0.09$ (step=$0.01$) and $0.1-0.9$ (step=$0.1$), which is used for plotting ROC curve. The parameters of the antagonistic center-surround spatial filter are set to $M_W=21, \sigma_W=1.5$.

\begin{figure*}[!htbp]
    \centering
    \subfigure[]{\label{fig:s04f10b}\includegraphics[width=0.3\textwidth]{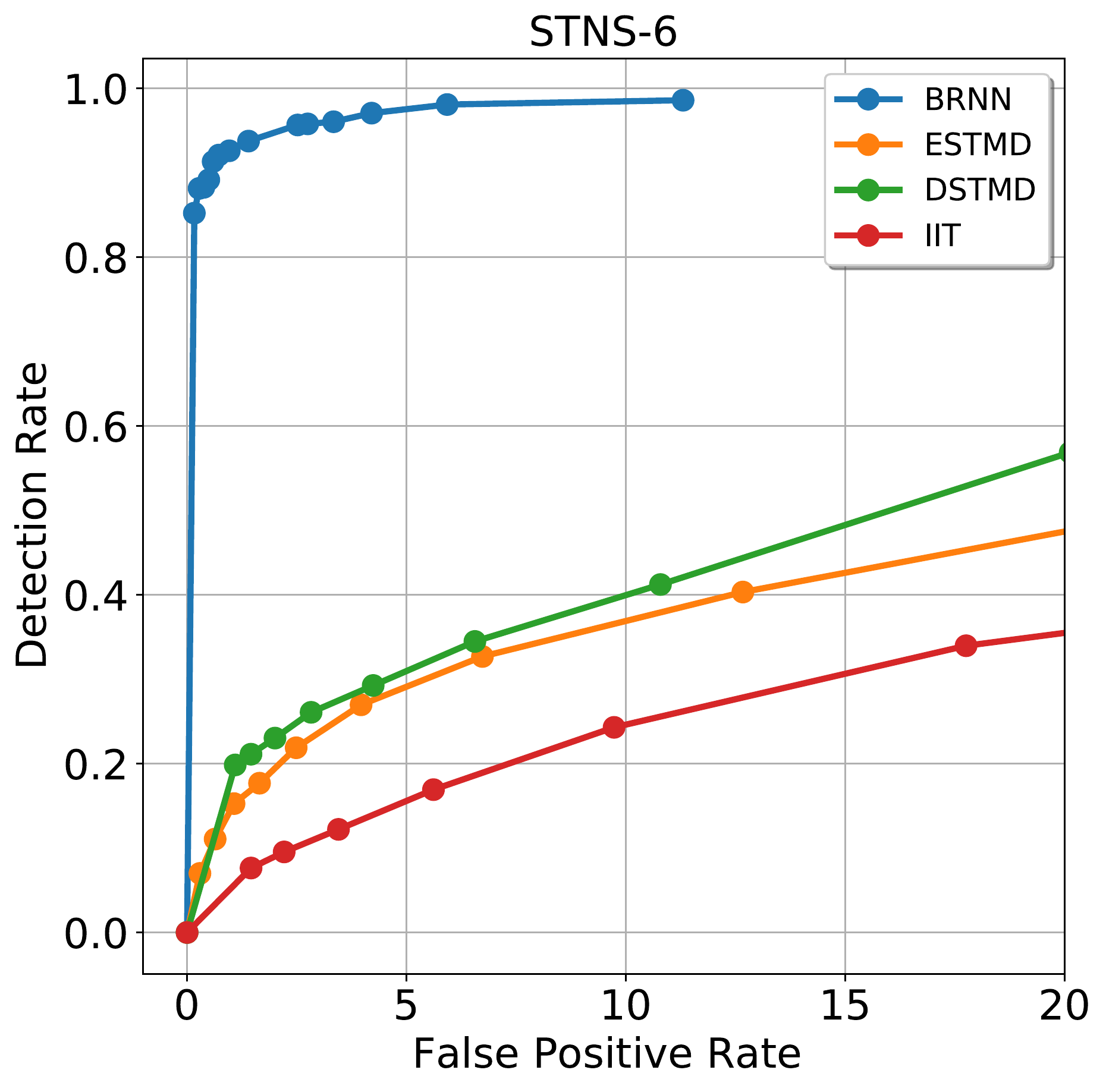}}
    ~\hspace{2pt}
    \subfigure[]{\label{fig:s04f10c}\includegraphics[width=0.3\textwidth]{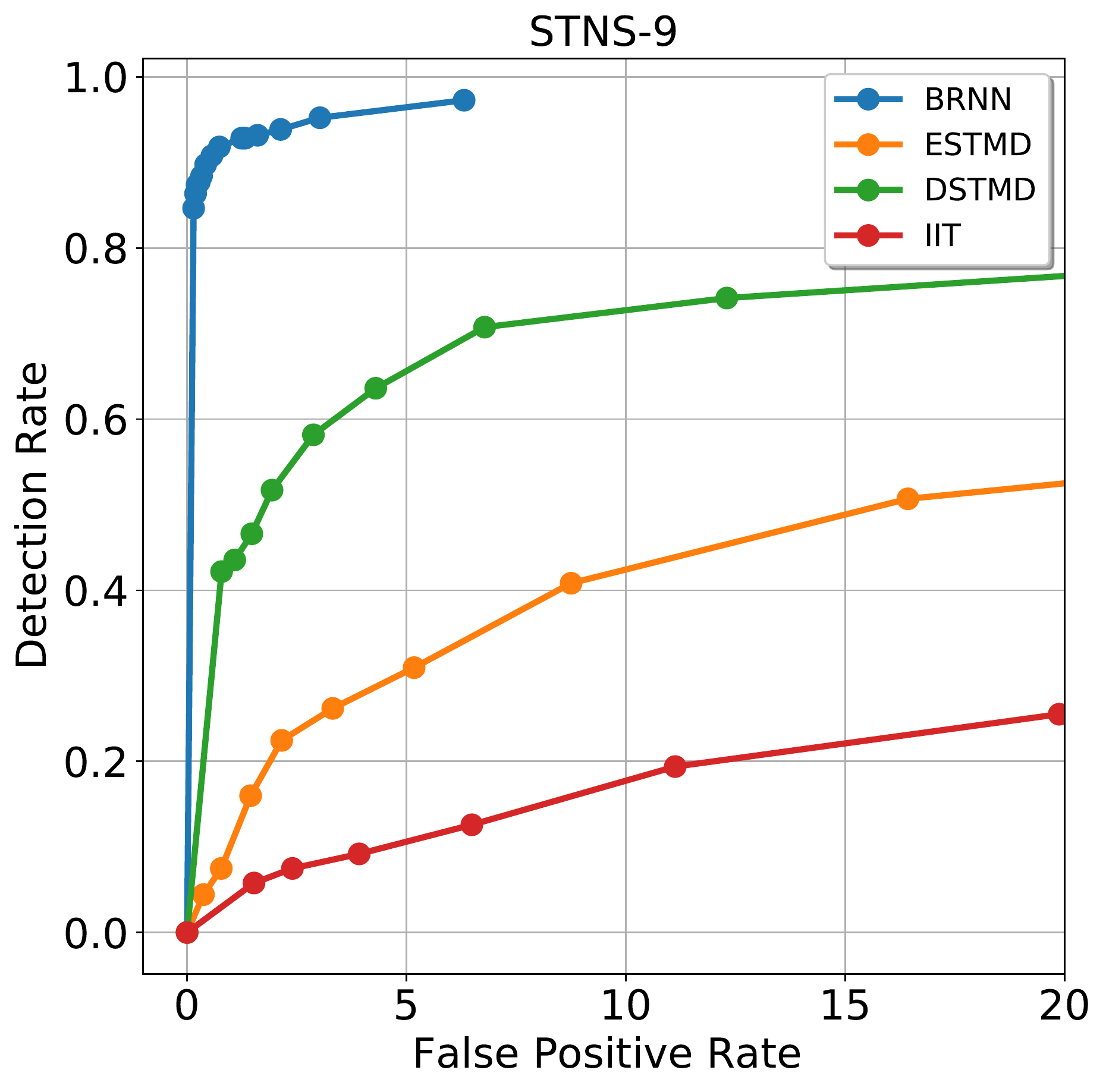}}
    ~\hspace{2pt}
    \subfigure[]{\label{fig:s04f10f}\includegraphics[width=0.3\textwidth]{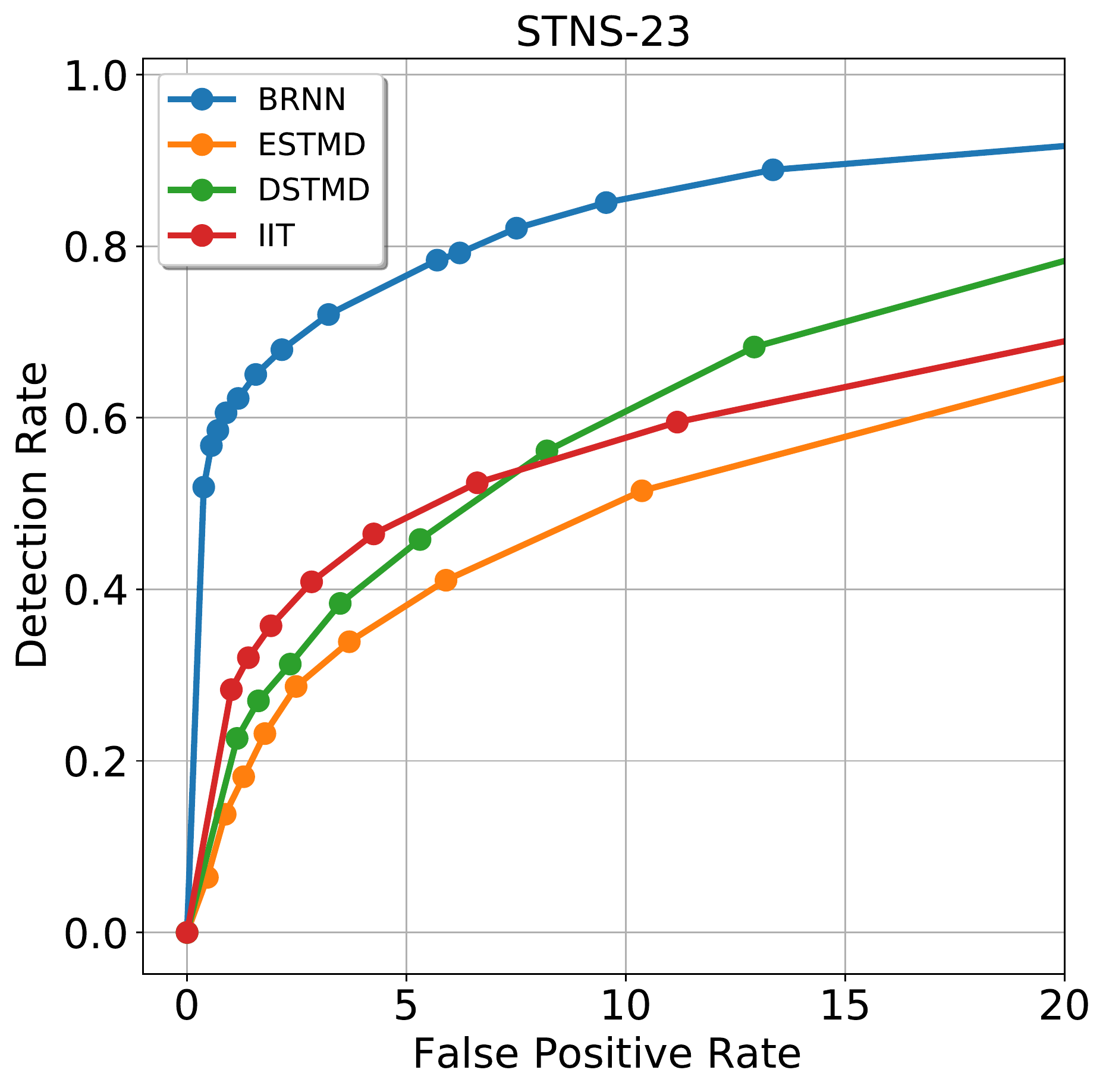}}
    \caption{The ROC curves of four bioinspired models for small-target motion detection in different STNS datasets.}
    \label{fig:s04f10}
\end{figure*}

We compare the results of ESTMD, DSTMD, IIT and our proposed BRNN at the same experimental conditions. The ROC curves of STNS-6, 9 and 23 are respectively plotted in Fig.\ref{fig:s04f10}, where the areas under the curve (AUC) of BRNN are clearly greater than other models. The ROC curves of other datasets are placed in the supplementary material. We list the detection rate of all scenarios at $FPR=5$ in Tab. 1. Obviously, the proposed model works accurately and stably for all different targets and cluttered backgrounds. 

\begin{table}[!htbp]
\caption{The detection rate of four models at $FPR=5$.} 
\centering 
\begin{tabular}{c c c c c c} 
\hline\hline 
Dataset & ESTMD & IIT & DSTMD & BRNN \\ [0.5ex] 
\hline 
STNS-1 & 0.11 & 0.16 & 0.34 & \textbf{0.66} \\ 
STNS-2 & 0.20 & 0.03 & 0.41 & \textbf{0.68} \\
STNS-3 & 0.10 & 0.12 & 0.16 & \textbf{0.96} \\
STNS-4 & 0.09 & 0.08 & 0.12 & \textbf{0.89} \\
STNS-5 & 0.12 & 0.07 & 0.48 & \textbf{0.97} \\
STNS-6 & 0.29 & 0.16 & 0.31 & \textbf{0.98} \\
STNS-7 & 0.06 & 0.06 & 0.12 & \textbf{0.88} \\
STNS-8 & 0.15 & 0.10 & 0.38 & \textbf{0.97} \\
STNS-9 & 0.31 & 0.11 & 0.66 & \textbf{0.96} \\
STNS-10 & 0.17 & 0.04 & 0.44 & \textbf{0.82} \\
STNS-11 & 0.09 & 0.08 & 0.37 & \textbf{0.95} \\
STNS-12 & 0.27 & 0.22 & 0.43 & \textbf{0.81} \\
STNS-13 & 0.31 & 0.29 & 0.46 & \textbf{0.48} \\
STNS-14 & 0.18 & 0.14 & 0.33 & \textbf{0.94} \\
STNS-15 & 0.24 & 0.11 & 0.28 & \textbf{0.74} \\
STNS-16 & 0.11 & 0.05 & 0.14 & \textbf{0.66} \\
STNS-17 & 0.17 & 0.08 & 0.11 & \textbf{0.64} \\
STNS-18 & 0.34 & 0.31 & 0.38 & \textbf{0.58} \\
STNS-19 & 0.21 & 0.19 & 0.26 & \textbf{0.84} \\
STNS-20 & 0.25 & 0.15 & 0.35 & \textbf{0.70} \\
STNS-21 & 0.15 & 0.23 & 0.27 & \textbf{0.74} \\
STNS-22 & 0.58 & 0.59 & \textbf{0.88} & 0.81 \\
STNS-23 & 0.38 & 0.48 & 0.54 & \textbf{0.77} \\
STNS-24 & 0.11 & 0.26 & 0.23 & \textbf{0.61} \\
STNS-25 & 0.21 & 0.20 & 0.38 & \textbf{0.91} \\ [1ex] 
\hline
Avg. & 0.21 & 0.17 & 0.35 & \textbf{0.80} \\ [1ex]
\hline
\end{tabular}
\label{tab:s04f01} 
\end{table}

\begin{table}[!htbp]
\caption{The computing time of four models for each step.} 
\centering 
\begin{tabular}{c c c c c} 
\hline\hline 
ESTMD & IIT & DSTMD & BRNN \\ [0.5ex] 
\hline 
$0.065\pm 0.004$ & $0.032 \pm 0.034$ & $0.085\pm 0.004$ & $0.084\pm 0.003$ \\ [1ex] 
\hline
\end{tabular}
\label{tab:s04f02} 
\end{table}

\begin{figure*}[!htbp]
    \centering
    \subfigure[]{
		\begin{minipage}[b]{0.25\linewidth}
		\includegraphics[width=1\linewidth]{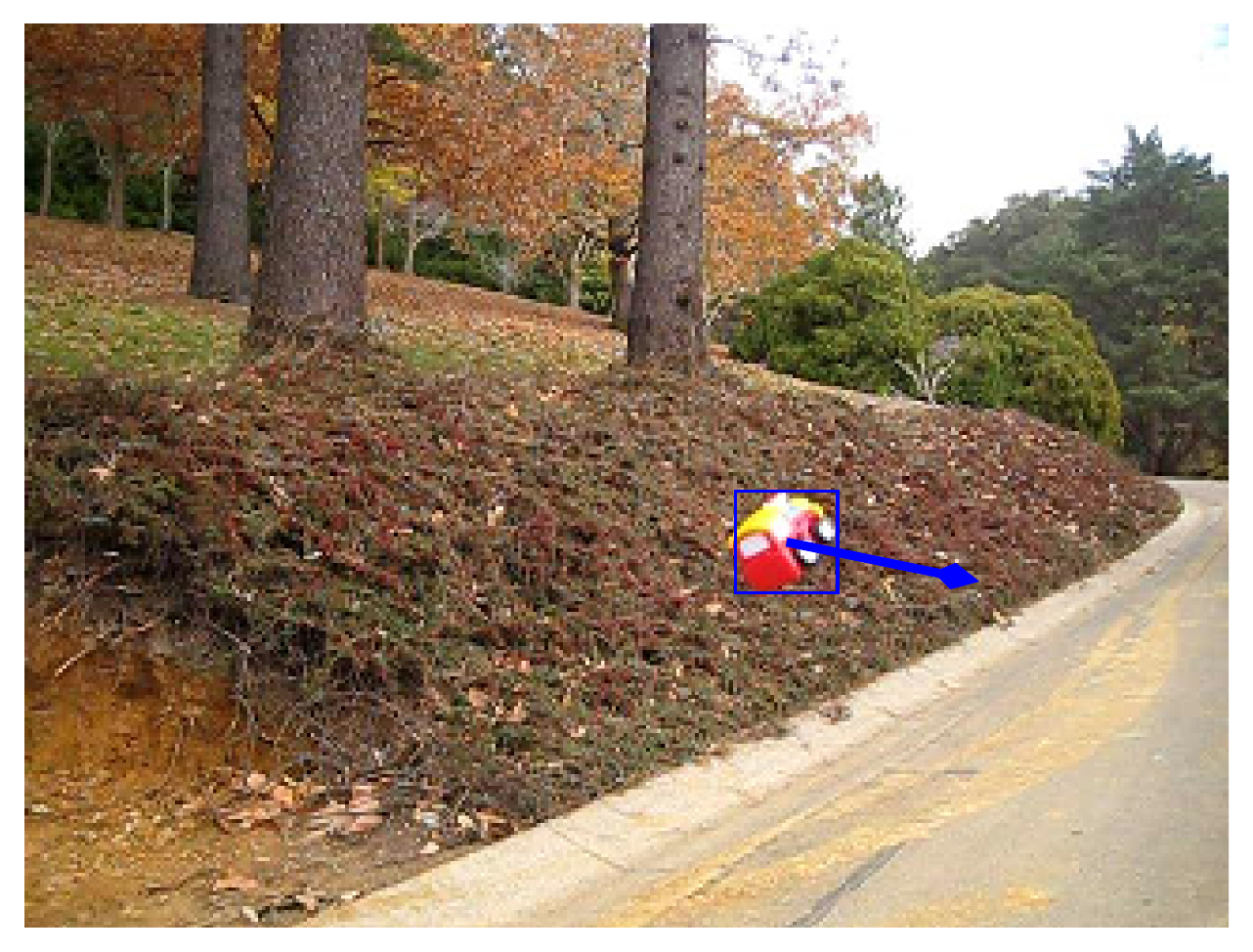}\vspace{1pt}
		\includegraphics[width=1\linewidth]{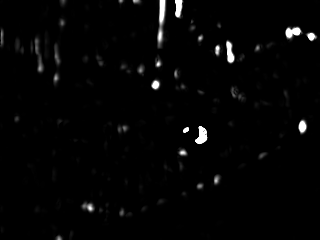}\vspace{2pt}
		\includegraphics[width=1\linewidth]{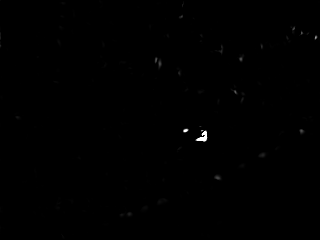}\vspace{2pt}
		\includegraphics[width=1\linewidth]{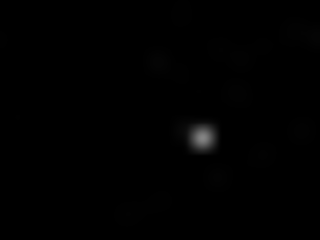}
		\end{minipage}}
	\subfigure[]{
		\begin{minipage}[b]{0.25\linewidth}
		\includegraphics[width=1\linewidth]{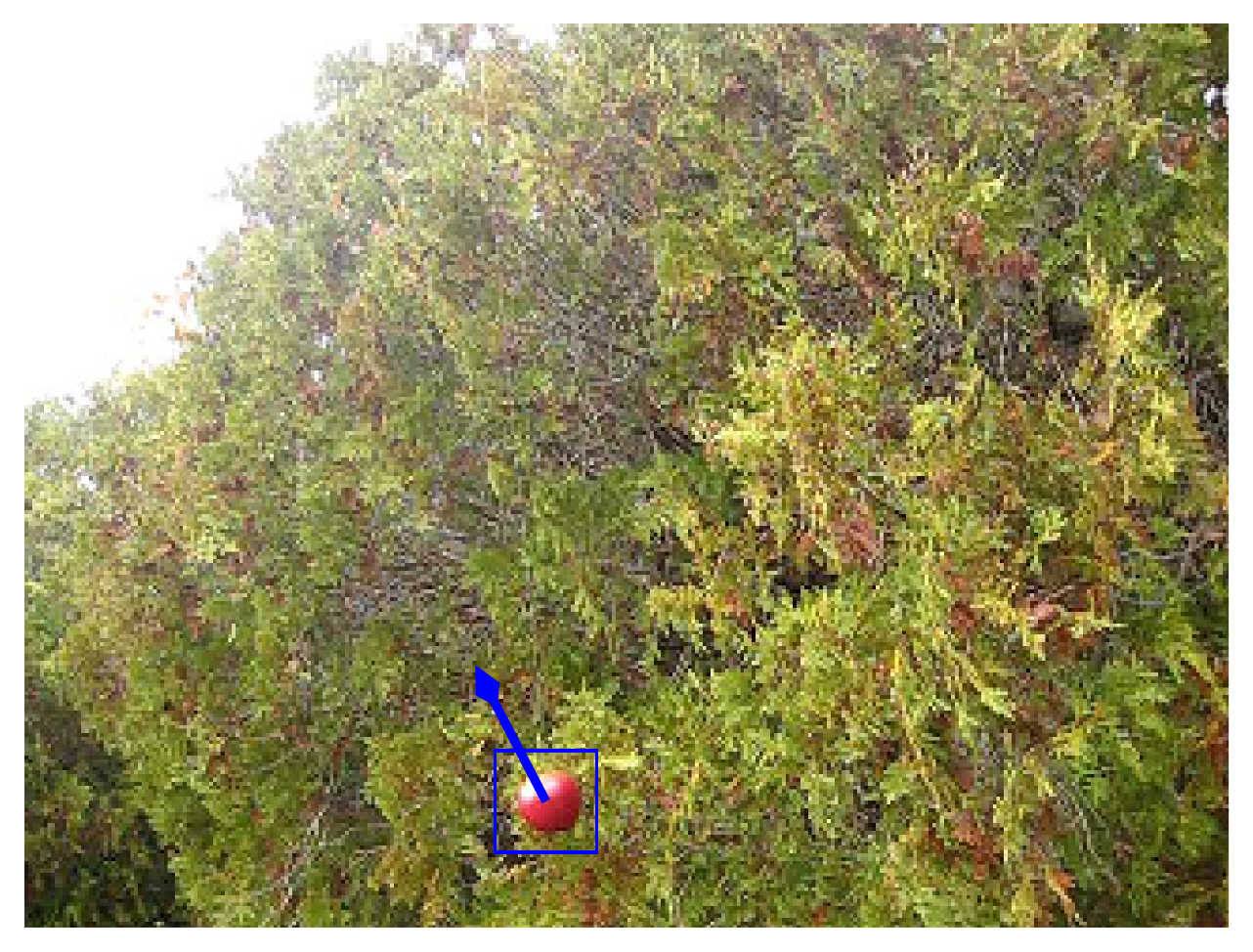}\vspace{1pt}
		\includegraphics[width=1\linewidth]{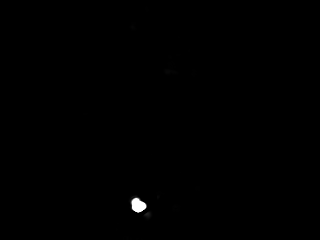}\vspace{2pt}
		\includegraphics[width=1\linewidth]{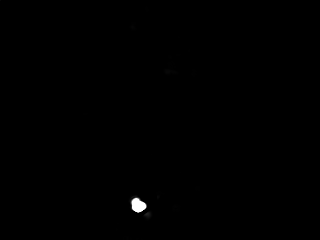}\vspace{2pt}
		\includegraphics[width=1\linewidth]{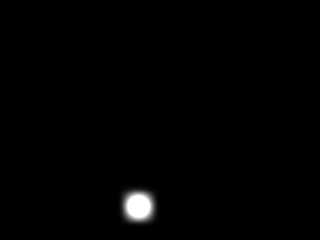}
		\end{minipage}}
	\subfigure[]{
		\begin{minipage}[b]{0.25\linewidth}
		\includegraphics[width=1\linewidth]{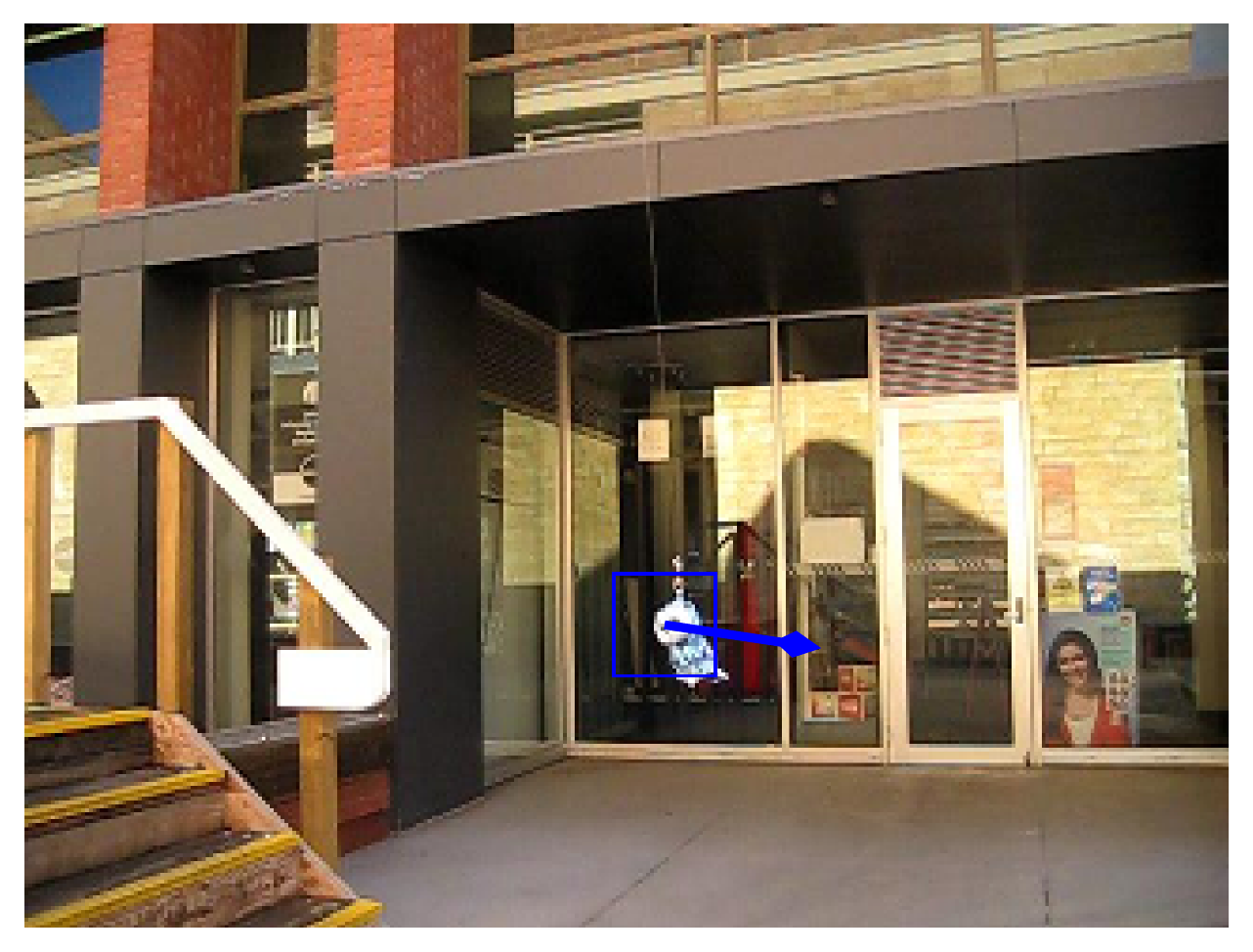}\vspace{1pt}
		\includegraphics[width=1\linewidth]{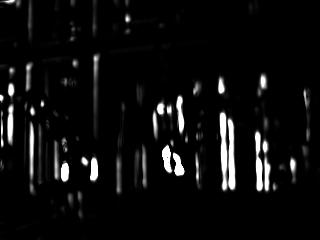}\vspace{2pt}
		\includegraphics[width=1\linewidth]{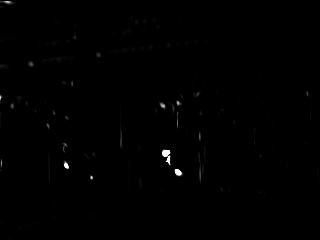}\vspace{2pt}
		\includegraphics[width=1\linewidth]{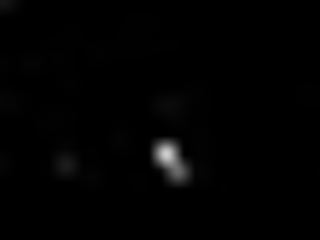}
		\end{minipage}}
    \caption{Neural responses of the proposed BRNN. Line 1: the position and motion direction of the small moving target. Line 2: the activities of DSGCs $V$. Line 3: the activities of DSGCs $V'$ with directionally selective inhibition. Line 4: The responses of $V'$ after Gaussian filtering.}
    \label{fig:s04f11}
\end{figure*}

This mainly benefits from two reasons. Firstly, the novel spatiotemporal energy model has a wider size and velocity sensitivity range such that it is effective to capture all possible targets. Secondly, this model can estimate motion direction accurately, which enables the directionally selective inhibition method to suppress the target-like features in the background successfully. Thus, the fake targets are removed effectively from the possible targets such that the detection rate increases and the false positives decreases. Taking STNS-6, 9 and 23 for example, we plot the activities of DSGCs $V$, DSGCs with directionally selective inhibition $V'$ and $V'$ after Gaussian filtering respectively. As shown in Fig. \ref{fig:s04f11}, the BRNN can not only accurately detect the position of small moving targets, but also extract their motion direction and motion energy at the same time. Meanwhile, the algorithm of directionally selective inhibition can suppress the activations caused by the moving background.

Finally, We further examine the processing speed of different models. All experiments are conducted in a computer, which has an ADM Ryzen 5 3600X 6-core 3.8GHz Processor and 16GB of RAM. It runs an Ubuntu 18.04 LTS 64-bit operating system. The architecture has been developed in Python 3.8 and OpenCV. The computing time is mainly related to the size of the input image and the complexity of model. With respect to the $320\times 240$ input image, the computing time for each step is recorded in Tab. \ref{tab:s04f02}. The IIT model has the fastest processing speed due to its relatively simple and optimized spatiotemporal filtering process. While the computing cost of the proposed BRNN is close to the DSTMD, since both of them use directionally selective spatial filtering to estimate motion direction and adopt the DBSCAN algorithm to select the region of interest.

\section{CONCLUSION}
It is significant and challenging for robotic visual systems to improve the diagnostic ability of small target motion detection in cluttered moving backgrounds. Inspired by elementary motion vision in the mammalian retina, this paper presented a bioinspired retinal neural network that can not only detect small moving targets in the cluttered moving background, but also extract the information of motion direction and motion energy. More specifically, there are two main contributions. Firstly, a novel spatiotemporal energy model has been proposed based on a neurodynamics-based temporal filtering and 2-D spatial Gabor filtering, simulating the basic functions of photoreceptor cells, bipolar cells, amacrine cells, and ganglion cells in the retina. It is able to estimate motion direction accurately via only two perpendicular spatiotemporal filtering signals, and respond to small targets of different sizes and velocities through changing the dendrite field size of spatial filter. Secondly, we further proposed an algorithm of directionally selective inhibition to suppress the target-like features in the moving background. This method is able to reduce the influence of background motion effectively. Comparing with other bioinspired models, the proposed model works stably for small targets of a wider size and velocity range, and has better detection performance in the synthetic and real data experiments. Meanwhile, it can also extract the information of motion direction and motion energy accurately and rapidly. The current research provides an attractive alternative paradigm for robotic search and tracking applications. It is also very meaningful for the research of biology and computational neuroscience.


%

%


\ifCLASSOPTIONcaptionsoff
  \newpage
\fi



\bibliographystyle{IEEEtran}
\bibliography{ref}

\begin{thebibliography}{10}
\providecommand{\url}[1]{#1}
\csname url@samestyle\endcsname
\providecommand{\newblock}{\relax}
\providecommand{\bibinfo}[2]{#2}
\providecommand{\BIBentrySTDinterwordspacing}{\spaceskip=0pt\relax}
\providecommand{\BIBentryALTinterwordstretchfactor}{4}
\providecommand{\BIBentryALTinterwordspacing}{\spaceskip=\fontdimen2\font plus
\BIBentryALTinterwordstretchfactor\fontdimen3\font minus
  \fontdimen4\font\relax}
\providecommand{\BIBforeignlanguage}[2]{{%
\expandafter\ifx\csname l@#1\endcsname\relax
\typeout{** WARNING: IEEEtran.bst: No hyphenation pattern has been}%
\typeout{** loaded for the language `#1'. Using the pattern for}%
\typeout{** the default language instead.}%
\else
\language=\csname l@#1\endcsname
\fi
#2}}
\providecommand{\BIBdecl}{\relax}
\BIBdecl

\bibitem{huang2013radial}
S.~Huang and B.~Do, ``Radial basis function based neural network for motion
  detection in dynamic scenes,'' \emph{IEEE transactions on cybernetics},
  vol.~44, no.~1, pp. 114--125, 2013.

\bibitem{yong2018robust}
H.~Yong, D.~Meng, W.~Zuo, and L.~Zhang, ``Robust online matrix factorization
  for dynamic background subtraction,'' \emph{IEEE Transactions on Pattern
  Analysis and Machine Intelligence}, vol.~40, no.~7, pp. 1726--1740, 2018.

\bibitem{li2016rotation}
Z.~Li, G.~Zhao, S.~Li, H.~Sun, R.~Tao, X.~Huang, and Y.~J. Guo, ``Rotation
  feature extraction for moving targets based on temporal differencing and
  image edge detection,'' \emph{IEEE Geoscience and Remote Sensing Letters},
  vol.~13, no.~10, pp. 1512--1516, 2016.

\bibitem{shuigen2009motion}
S.~Wei, C.~Zhen, and H.~Dong, ``Motion detection based on temporal difference
  method and optical flow field,'' in \emph{International Symposium on
  Electronic Commerce and Security}, vol.~2, 2009, pp. 85--88.

\bibitem{tsai2009independent}
D.~Tsai and S.~Lai, ``Independent component analysis-based background
  subtraction for indoor surveillance,'' \emph{IEEE Transactions on Image
  Processing}, vol.~18, no.~1, pp. 158--167, 2009.

\bibitem{woo2010environmentally}
H.~Woo, Y.~M. Jung, J.~Kim, and J.~K. Seo, ``Environmentally robust motion
  detection for video surveillance,'' \emph{IEEE Transactions on Image
  Processing}, vol.~19, no.~11, pp. 2838--2848, 2010.

\bibitem{fortun2015optical}
D.~Fortun, P.~Bouthemy, and C.~Kervrann, ``Optical flow modeling and
  computation,'' \emph{Computer Vision and Image Understanding}, vol. 134, pp.
  1--21, 2015.

\bibitem{wei2011motion}
S.~Wei, L.~Yang, Z.~Chen, and Z.~Liu, ``Motion detection based on optical flow
  and self-adaptive threshold segmentation,'' \emph{Procedia Engineering},
  vol.~15, pp. 3471--3476, 2011.

\bibitem{girshick2016region}
R.~Girshick, J.~Donahue, T.~Darrell, and J.~Malik, ``Region-based convolutional
  networks for accurate object detection and segmentation,'' \emph{IEEE
  transactions on pattern analysis and machine intelligence}, vol.~38, no.~1,
  p. 142–158, Jan. 2016.

\bibitem{girshick2015fast}
R.~Girshick, ``Fast r-cnn,'' in \emph{Proceedings of the IEEE international
  conference on computer vision}, 2015, pp. 1440--1448.

\bibitem{ren2015faster}
S.~Ren, K.~He, R.~Girshick, and J.~Sun, ``Faster r-cnn: Towards real-time
  object detection with region proposal networks,'' in \emph{Advances in neural
  information processing systems}, 2015, pp. 91--99.

\bibitem{liu2016ssd}
W.~Liu, D.~Anguelov, D.~Erhan, C.~Szegedy, S.~Reed, C.~Y. Fu, and A.~C. Berg,
  ``Ssd: Single shot multibox detector,'' in \emph{European conference on
  computer vision}.\hskip 1em plus 0.5em minus 0.4em\relax Springer, 2016, pp.
  21--37.

\bibitem{redmon2018yolov3}
J.~Redmon and A.~Farhadi, ``Yolov3: An incremental improvement,'' \emph{arXiv
  preprint arXiv:1804.02767}, 2018.

\bibitem{redmon2017yolo9000}
J.~Redmon and A.~Farhadi, ``Yolo9000: better, faster, stronger,'' in
  \emph{Proceedings of the IEEE conference on computer vision and pattern
  recognition}, 2017, pp. 7263--7271.

\bibitem{redmon2016you}
J.~Redmon, S.~Divvala, R.~Girshick, and A.~Farhadi, ``You only look once:
  Unified, real-time object detection,'' in \emph{Proceedings of the IEEE
  conference on computer vision and pattern recognition}, 2016, pp. 779--788.

\bibitem{gao2013infrared}
C.~Gao, D.~Meng, Y.~Yang, Y.~Wang, X.~Zhou, and A.~G. Hauptmann, ``Infrared
  patch-image model for small target detection in a single image,'' \emph{IEEE
  Transactions on Image Processing}, vol.~22, no.~12, pp. 4996--5009, 2013.

\bibitem{bai2018derivative}
X.~Bai and Y.~Bi, ``Derivative entropy-based contrast measure for infrared
  small-target detection,'' \emph{IEEE Transactions on Geoscience and Remote
  Sensing}, vol.~56, no.~4, pp. 2452--2466, 2018.

\bibitem{dong2014a}
X.~Dong, X.~Huang, Y.~Zheng, S.~Bai, and W.~Xu, ``A novel infrared small moving
  target detection method based on tracking interest points under complicated
  background,'' \emph{Infrared Physics \& Technology}, vol.~65, pp. 36--42,
  2014.

\bibitem{lin2018using}
L.~Lin, S.~Wang, and Z.~Tang, ``Using deep learning to detect small targets in
  infrared oversampling images,'' \emph{Journal of Systems Engineering and
  Electronics}, vol.~29, no.~5, pp. 947--952, 2018.

\bibitem{sun2018sg}
M.~Sun, Z.~Zhou, Q.~Hu, Z.~Wang, and J.~Jiang, ``Sg-fcn: A motion and
  memory-based deep learning model for video saliency detection,'' \emph{IEEE
  transactions on cybernetics}, vol.~49, no.~8, pp. 2900--2911, 2018.

\bibitem{zhu2020moving}
J.~Zhu, Z.~Wang, S.~Wang, and S.~Chen, ``Moving object detection based on
  background compensation and deep learning,'' \emph{Symmetry}, vol.~12,
  no.~12, p. 1965, 2020.

\bibitem{nordstrom2006small}
K.~Nordstr{\"o}m and D.~C. O'Carroll, ``Small object detection neurons in
  female hoverflies,'' \emph{Proceedings of the Royal Society B: Biological
  Sciences}, vol. 273, no. 1591, pp. 1211--1216, 2006.

\bibitem{barnett2007retinotopic}
P.~D. Barnett, K.~Nordstr{\"o}m, and D.~C. O'Carroll, ``Retinotopic
  organization of small-field-target-detecting neurons in the insect visual
  system,'' \emph{Current Biology}, vol.~17, no.~7, pp. 569--578, 2007.

\bibitem{wiederman2008model}
S.~D. Wiederman, P.~A. Shoemaker, and D.~C. O'Carroll, ``A model for the
  detection of moving targets in visual clutter inspired by insect
  physiology,'' \emph{PloS one}, vol.~3, no.~7, p. e2784, 2008.

\bibitem{bagheri2017autonomous}
Z.~M. Bagheri, B.~S. Cazzolato, S.~Grainger, D.~C. O’Carroll, and S.~D.
  Wiederman, ``An autonomous robot inspired by insect neurophysiology pursues
  moving features in natural environments,'' \emph{Journal of neural
  engineering}, vol.~14, no.~4, p. 046030, 2017.

\bibitem{bagheri2017performance}
Z.~M. Bagheri, S.~D. Wiederman, B.~S. Cazzolato, S.~Grainger, and D.~C.
  O’Carroll, ``Performance of an insect-inspired target tracker in natural
  conditions,'' \emph{Bioinspiration \& biomimetics}, vol.~12, no.~2, p.
  025006, 2017.

\bibitem{bagheri2015properties}
Z.~M. Bagheri, S.~D. Wiederman, B.~S. Cazzolato, S.~Grainger, and D.~C.
  O'Carroll, ``Properties of neuronal facilitation that improve target tracking
  in natural pursuit simulations,'' \emph{Journal of The Royal Society
  Interface}, vol.~12, no. 108, p. 20150083, 2015.

\bibitem{wang2020a}
H.~Wang, J.~Peng, and S.~Yue, ``A directionally selective small target motion
  detecting visual neural network in cluttered backgrounds,'' \emph{IEEE
  Transactions on Cybernetics}, vol.~50, no.~4, pp. 1541--1555, 2020.

\bibitem{wang2020arobust}
H.~Wang, J.~Peng, X.~Zheng, and S.~Yue, ``A robust visual system for small
  target motion detection against cluttered moving backgrounds,'' \emph{IEEE
  Transactions on Neural Networks and Learning Systems}, vol.~31, no.~3, pp.
  839--853, 2020.

\bibitem{colonnier2019bio}
F.~Colonnier, S.~Ramirez-Martinez, S.~Viollet, and F.~Ruffier, ``A bio-inspired
  sighted robot chases like a hoverfly,'' \emph{Bioinspiration \& biomimetics},
  vol.~14, no.~3, p. 036002, 2019.

\bibitem{caves2018visual}
E.~M. Caves, N.~C. Brandley, and S.~Johnsen, ``Visual acuity and the evolution
  of signals,'' \emph{Trends in ecology \& evolution}, vol.~33, no.~5, pp.
  358--372, 2018.

\bibitem{bostrom2016ultra}
J.~E. Bostr{\"o}m, M.~Dimitrova, C.~Canton, O.~H{\aa}stad, A.~Qvarnstr{\"o}m,
  and A.~{\"O}deen, ``Ultra-rapid vision in birds,'' \emph{PLoS One}, vol.~11,
  no.~3, p. e0151099, 2016.

\bibitem{borst2015common}
A.~Borst and M.~Helmstaedter, ``Common circuit design in fly and mammalian
  motion vision,'' \emph{Nature Neuroscience}, vol.~18, no.~8, pp. 1067--1076,
  2015.

\bibitem{clark2016parallel}
D.~Clark and J.~B. Demb, ``Parallel computations in insect and mammalian visual
  motion processing,'' \emph{Current Biology}, vol.~26, no.~20, 2016.

\bibitem{adelson1985spatiotemporal}
E.~H. Adelson and J.~R. Bergen, ``Spatiotemporal energy models for the
  perception of motion,'' \emph{Journal of The Optical Society of America
  A-optics Image Science and Vision}, vol.~2, no.~2, pp. 284--299, 1985.

\bibitem{heeger1988optical}
D.~J. Heeger, ``Optical flow using spatiotemporal filters,''
  \emph{International journal of computer vision}, vol.~1, no.~4, pp. 279--302,
  1988.

\bibitem{browning2009a}
N.~A. Browning, S.~Grossberg, and E.~Mingolla, ``A neural model of how the
  brain computes heading from optic flow in realistic scenes.'' \emph{Cognitive
  Psychology}, vol.~59, no.~4, pp. 320--356, 2009.

\bibitem{browning2009cortical}
N.~A. Browning, S.~Grossberg, and E.~Mingolla, ``Cortical dynamics of
  navigation and steering in natural scenes: Motion-based object segmentation,
  heading, and obstacle avoidance,'' \emph{Neural Networks}, vol.~22, no.~10,
  pp. 1383--1398, 2009.

\bibitem{masland2001fundamental}
R.~H. Masland, ``The fundamental plan of the retina,'' \emph{Nature
  neuroscience}, vol.~4, no.~9, pp. 877--886, 2001.

\bibitem{kendel2000principles}
E.~R. Kandel, J.~H. Schwartz, T.~M. Jessell, S.~A. Siegelbaum, and H.~A. J.,
  \emph{Principles of Neural Science}.\hskip 1em plus 0.5em minus 0.4em\relax
  New York: McGraw-hill, 2000, vol.~4.

\bibitem{euler2014retinal}
T.~Euler, S.~Haverkamp, T.~Schubert, and T.~Baden, ``Retinal bipolar cells:
  elementary building blocks of vision,'' \emph{Nature Reviews Neuroscience},
  vol.~15, no.~8, pp. 507--519, 2014.

\bibitem{bialek1990temporal}
W.~Bialek and W.~Owen, ``Temporal filtering in retinal bipolar cells. elements
  of an optimal computation?'' \emph{Biophysical journal}, vol.~58, no.~5, pp.
  1227--1233, 1990.

\bibitem{burkhardt2007retinal}
D.~A. Burkhardt, P.~K. Fahey, and M.~A. Sikora, ``Retinal bipolar cells :
  Temporal filtering of signals from cone photoreceptors,'' \emph{Visual
  Neuroscience}, vol.~24, no.~6, pp. 765--774, 2007.

\bibitem{Hoggarth2015}
A.~Hoggarth, A.~J. McLaughlin, K.~Ronellenfitch, S.~Trenholm, R.~Vasandani,
  S.~Sethuramanujam, D.~Schwab, K.~L. Briggman, and G.~B. Awatramani,
  ``Specific wiring of distinct amacrine cells in the directionally selective
  retinal circuit permits independent coding of direction and size,''
  \emph{Neuron}, vol.~86, no.~1, pp. 276--291, 2015.

\bibitem{dhande2015contributions}
O.~S. Dhande, B.~K. Stafford, J.-H.~A. Lim, and A.~D. Huberman, ``Contributions
  of retinal ganglion cells to subcortical visual processing and behaviors,''
  \emph{Annual review of vision science}, vol.~1, pp. 291--328, 2015.

\bibitem{fu2018shaping}
Q.~Fu, C.~Hu, J.~Peng, and S.~Yue, ``Shaping the collision selectivity in a
  looming sensitive neuron model with parallel on and off pathways and spike
  frequency adaptation,'' \emph{Neural Networks}, vol. 106, pp. 127--143, 2018.

\bibitem{werner1961autocorrelation}
R.~Werner, ``Autocorrelation, a principle for the evaluation of sensory
  information by the central nervous system,'' \emph{Sensory Communication. The
  MIT Press}, 1961.

\bibitem{ester1996density}
M.~Ester, H.~P. Kriegel, J.~Sander, X.~Xu \emph{et~al.}, ``A density-based
  algorithm for discovering clusters in large spatial databases with noise.''
  in \emph{Kdd}, vol.~96, no.~34, 1996, pp. 226--231.

\bibitem{schubert2017dbscan}
E.~Schubert, J.~Sander, M.~Ester, H.~P. Kriegel, and X.~Xu, ``Dbscan revisited,
  revisited: why and how you should (still) use dbscan,'' \emph{ACM
  Transactions on Database Systems (TODS)}, vol.~42, no.~3, pp. 1--21, 2017.

\bibitem{wyatt1975directionally}
H.~J. Wyatt and N.~W. Daw, ``Directionally sensitive ganglion cells in the
  rabbit retina: specificity for stimulus direction, size, and speed,''
  \emph{Journal of Neurophysiology}, vol.~38, no.~3, pp. 613--626, 1975.

\end{thebibliography}
%
%





\end{document}